\documentclass[letterpaper]{article} 
\usepackage{aaai2026}
\nocopyright
\usepackage{times}  
\usepackage{helvet}  
\usepackage{courier}  
\usepackage[hyphens]{url}  
\usepackage{graphicx} 
\usepackage{amsmath}
\usepackage{amssymb}
\usepackage[dvipsnames]{xcolor}
\usepackage{natbib}  
\usepackage{caption} 
\usepackage{algorithm}
\usepackage{multirow}
\usepackage{siunitx}
\usepackage{booktabs}
\usepackage{algpseudocode}

\usepackage{amsthm} 
\newtheorem{lemma}{Lemma}
\usepackage{float}
\usepackage{newfloat}
\usepackage{listings}
\DeclareCaptionStyle{ruled}{labelfont=normalfont,labelsep=colon,strut=off} 
\floatstyle{ruled}
\newfloat{listing}{tb}{lst}{}
\floatname{listing}{Listing}
\title{Beyond the Mean: Fisher-Orthogonal Projection for Natural Gradient Descent in Large Batch Training}

\author{
    Yishun Lu \textsuperscript{\rm 1}, Wesley Armour \textsuperscript{\rm 1}
}

\affiliations{
    \normalfont \fontsize{10}{12}\selectfont
    \textsuperscript{1}Oxford e-Research Centre, Department of Engineering Science, University of Oxford\\
    yishun.lu@eng.ox.ac.uk, wes.armour@oerc.ox.ac.uk
}

\usepackage{bibentry}

\begin{document}

\maketitle

\begin{abstract}

Modern GPUs are equipped with large amounts of high-bandwidth memory, enabling them to support mini-batch sizes of up to tens of thousands of training samples. However, most existing optimizers struggle to perform effectively at such a large batch size. 
As batch size increases, gradient noise decreases due to averaging over many samples, limiting the ability of first-order methods to escape sharp or suboptimal minima and reach the global minimum.
Meanwhile, second-order methods like the natural gradient with Kronecker-Factored Approximate Curvature (KFAC) often require excessively high damping to remain stable at large batch sizes. This high damping effectively ``washes out" the curvature information that gives these methods their advantage, reducing their performance to that of simple gradient descent.
In this paper, we introduce Fisher-Orthogonal Projection (FOP), a novel technique that restores the effectiveness of the second-order method at very large batch sizes, enabling scalable training with improved generalization and faster convergence.
FOP constructs a variance-aware update direction by leveraging gradients from two sub-batches, enhancing the average gradient with a component of the gradient difference that is orthogonal to the average under the Fisher-metric. 
Through extensive benchmarks, we show that FOP accelerates convergence by $\times1.2–1.3$ over KFAC and $\times1.5–1.7$ over SGD/AdamW at the same moderate batch sizes, while at extreme scales it achieves up to a $\times7.5$ speedup. 
Unlike other methods, FOP maintains small-batch accuracy when scaling to extremely large batch sizes. Moreover, it reduces Top-1 error by 2.3–3.3\% on long-tailed CIFAR benchmarks, demonstrating robust generalization under severe class imbalance. Our lightweight, geometry-aware use of intra-batch variance makes natural-gradient optimization practical on modern data-centre GPUs. FOP is open-source and pip-installable, which can be integrated into existing training code with a single line and no extra configuration.

\end{abstract}

\begin{links}
    \link{Code}{https://github.com/yishunlu-222/fop.git}
    \link{Extended version}{https://arxiv.org/abs/2508.13898}
\end{links}

\section{Introduction}
The increasing scale of modern language models and vision transformers has made large mini-batch training a necessity. Modern GPUs offer extensive high-bandwidth memory (such as 192GB in AMD MI300X), and data centers often combine hundreds of these devices, enabling the efficient processing of tens of thousands of training examples per batch. This improves hardware utilization, reduces communication overhead, and accelerates training significantly.
\noindent
As batch sizes increase, the gradients become more deterministic. This reduces the stochastic noise that previously helped first-order optimizers such as SGD~\cite{sgd}, Adam~\cite{adam}, and AdamW~\cite{adamw} to explore flatter minima of the loss landscape. The loss of this noise ~\cite{keskar2016large,mccandlish2018empirical,you2019lamb} requires first-order methods to rely on smaller learning rates and stronger explicit regularization to maintain stability and generalization performance.


\noindent
\noindent
Natural-Gradient Descent (NGD) addresses scenarios where stochastic gradient noise is minimal, such as very large mini-batches~\cite{pascanu2013revisiting,shazeer2018adafactor,ishikawa2024does}. Unlike first-order methods, NGD incorporates second-order curvature information via the Fisher information matrix~\cite{amari1998natural}, enabling geometry-aware parameter updates invariant to model parameterization. However, exact computation of the Fisher matrix is infeasible for modern neural networks. Practical implementations rely on approximations, with Kronecker-Factored Approximate Curvature (KFAC) being the most widely adopted~\cite{martens2015optimizing}. KFAC simplifies the Fisher matrix by exploiting the layer-wise structure of neural networks, making the approximation computationally efficient. Subsequent methods further approximate KFAC to improve tractability~\cite{lin2024structured,yang2020sketchy,liu2024layer,eschenhagen2023kronecker}.
Despite its promise, KFAC struggles in the very-large-batch regime required for modern hardware. As the batch size grows, the Fisher matrix becomes increasingly ill-conditioned \cite{sagun2016eigenvalues,ghorbani2019investigation}, leading to numerical instability. This forces the use of strong damping, which unfortunately suppresses the very curvature information that gives KFAC its advantage. 

\noindent
Prior attempts to scale natural-gradient methods to large-batch training include SENG~\citep{yang2020sketchy}, which uses low-rank sketches but introducing new hyperparameters, and SP-NGD~\citep{osawa2020scalable}, which relies on empirical-Fisher approximations and stale-statistic heuristics that require task-specific hyperparameter retuning. Adaptive batch-size schedules~\citep{yao2018large,shi2023distributedshampoo} improve throughput but still rely on heavily damped mean gradients. Although these methods reduce hardware requirements, their update rules remain fundamentally unchanged, still dominated by mean gradients and extensive hyperparameter tuning.


\noindent
In this paper, we propose Fisher-Orthogonal Projection (FOP), which augments natural gradient descent with a Fisher-orthogonal variance correction. This novel geometry-aware update captures intra-batch gradient variation without relying on sketching ranks, stale-statistics heuristics, or customized communication strategies. Specifically, our contributions include:
\begin{itemize}
  \item \textbf{Fisher–Orthogonal Projection optimizer.} We propose a novel second‐order update that augments the natural gradient with a geometry‐aware, variance‐controlled component, capturing intra‐batch information by standard KFAC.
  \item \textbf{Extreme large‐batch scalability.} We demonstrate that FOP seamlessly scales to cases where SGD, AdamW, and K‐FAC break down, and it can achieve speedups of up to $\times7.5$ in wall‐clock time while maintaining convergence at extremely large batch sizes in ImageNet and CIFAR datasets.
  \item \textbf{Robust generalization under imbalance.} Reducing Top-1 error rate by 2.3–3.3\% on long-tailed CIFAR-LT benchmarks without additional tricks
\item \textbf{Distributed FOP.} Efficiently sharding Fisher computation across GPUs with dual-gradient AllReduce and broadcasted updates, enabling scalable, low-overhead training.

\end{itemize}



\section{Natural Gradient Descent}

Natural Gradient Descent is a second-order optimization method derived from Newton's method, where the Hessian is replaced by the Fisher Information Matrix to reflect the geometry of the parameter space. In standard gradient descent, the update rule is:
\[
\theta_i = \theta_{i-1} - \eta \nabla_\theta \mathcal{L}(\theta_{i-1})
\]
where \( \theta_i \) is the parameter vector at iteration \( i \), \( \eta \) is the learning rate, and \( \nabla_\theta \mathcal{L}(\theta_{i-1}) \) is the gradient of the loss function \( \mathcal{L} \) evaluated at \( \theta_{i-1} \).
This treats all directions in parameter space uniformly, which can lead to slow convergence in ill-conditioned landscapes. Newton's method improves this by preconditioning the gradient with the inverse Hessian \( H^{-1} \), but computing the full Hessian is often infeasible for large models.

\noindent
Natural Gradient Descent modifies the update by replacing the Hessian with the Fisher Information Matrix \( F \), which defines a Riemannian metric on the statistical manifold \cite{amari1998natural}:
\[
\theta_i = \theta_{i-1} - \eta F^{-1} \nabla_\theta \mathcal{L}(\theta_{i-1})
\]
This yields the steepest descent direction under the Kullback–Leibler (KL) divergence, making the update invariant to parameter reparameterizations ~\cite{martens2015optimizing}.

\section{Fisher‐Orthogonal Projection}

\begin{figure}
    \centering
    \includegraphics[width=1\linewidth]{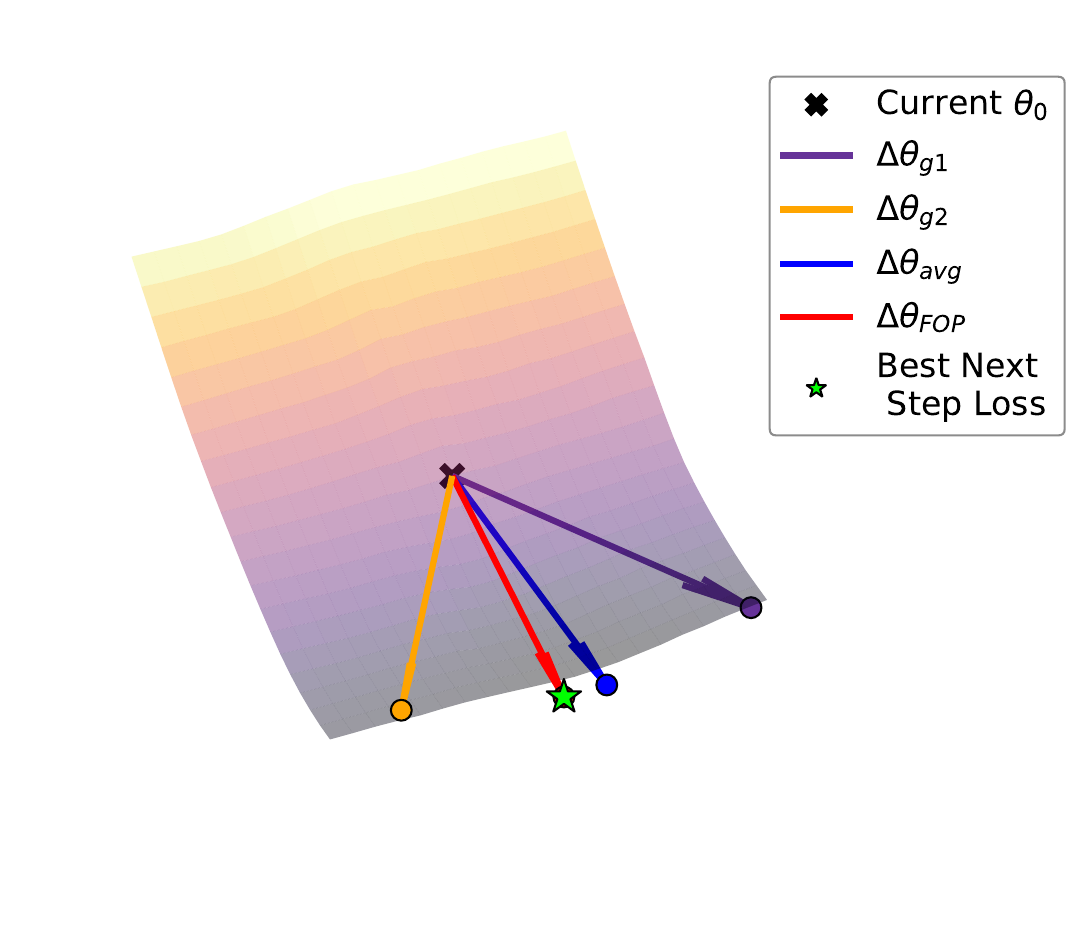}
    \caption{3D loss landscape of training ResNet-18 with CIFAR-10 for batch size of 1024. This visualization relies on the method suggested in~\cite{li2018visualizing}. Arrows represent the direction of the steps of different gradients. The green star is the smallest loss after updating the model based on different update directions. }
    \label{fig:loss landscape}
\end{figure}

\noindent
A loss landscape is plotted based on the method suggested in~\cite{li2018visualizing} in Figure~\ref{fig:loss landscape}, by projecting the high-dimensional parameter space onto a 2D plane using two random but normalized directions and evaluating the loss across a grid in this plane. Naively averaging gradients across mini-batches can obscure useful optimization directions due to averaging over many samples. In particular, such averaging may suppress informative signals when gradients from different batches point in significantly different directions. Especially at large batch sizes, where gradient variability is low but intra-batch differences can still carry important optimization signals.
We propose the \textit{Fisher-Orthogonal Projection (FOP)} method, which preserves the informative structure of each mini-batch gradient while ensuring stable and curvature-aware descent. The key idea is to use the average gradient as the primary descent direction, capturing the common signal shared across mini-batches. In addition, FOP introduces a Fisher-orthogonal component, derived from the difference between two mini-batch gradients. This orthogonal component contains complementary curvature-sensitive information that would otherwise be lost through simple averaging.

\noindent Suppose that at a parameter point \( \theta \), we compute two gradients \( g_1 \) and \( g_2 \) from two independent mini-batches. Then the FOP update is defined as:


\begin{equation}
L_1(\theta), \quad L_2(\theta)
\label{eq:batch_losses}
\end{equation}

\noindent
each associated with a gradient:

\begin{equation}
g_1 = \nabla_\theta L_1(\theta), \quad g_2 = \nabla_\theta L_2(\theta)
\label{eq:batch_gradients}
\end{equation}






\noindent
We compute the average and difference of the two gradients:

\begin{equation}
g_{\text{avg}} = \frac{1}{2}(g_{\text{1}} + g_{\text{2}}), \quad g_{\text{diff}} = g_{\text{1}} - g_{\text{2}}.
\label{eq:avg_gradient}
\end{equation}



\noindent
To eliminate redundancy and extract only novel information, we orthogonalize \( g_{\text{diff}} \) with respect to \( g_{\text{avg}} \) under the inner product induced by the Fisher matrix \( F \). The projection scalar is computed as:

\begin{equation}
s_{\text{proj}} = \frac{g_{\text{diff}}^\top F g_{\text{avg}}}{g_{\text{avg}}^\top F g_{\text{avg}} + \epsilon}
\label{eq:projection_scalar}
\end{equation}


\noindent
and the orthogonal component is then:

\begin{equation}
g_{\text{diff}}^\perp = g_{\text{diff}} - s_{\text{proj}} \cdot g_{\text{avg}}.
\label{eq:orthogonal_component}
\end{equation}

\noindent
By construction, this ensures that \( \langle g_{\text{avg}}, g_{\text{diff}}^\perp \rangle_F = 0 \) (as described in Lemma 1 in the Supplementary material in the extended version), meaning the new component contains only the information that is orthogonal in the Fisher geometry, which is already captured in the average gradient.

\noindent
The final combined update direction is given by:

\begin{equation}
g_{\text{combined}} = g_{\text{avg}} + \beta g_{\text{diff}}^\perp,
\label{eq:combined_gradient}
\end{equation}

\noindent
where \( \beta \) is a scalar weight, adaptively determined to locally minimize the primary or total loss. The overall parameter update using Natural Gradient Descent then becomes:

\begin{equation}
\theta_{t+1} = \theta_t - \eta F^{-1} g_{\text{combined}}.
\label{eq:ngd_update}
\end{equation}


\subsection{Layer-wise Adaptive Coefficient \( \beta \)}

\noindent
When the true optimization objective is the sum of two per-batch losses, we can write the total loss as:
\begin{equation}
L_{\text{tot}}(\theta) = L_1(\theta) + L_2(\theta).
\label{eq:total_loss}
\end{equation}

\noindent
To minimize this objective locally, we seek an optimal mixing coefficient \( \beta \) that balances the direction \( g_{\text{diff}}^\perp \) relative to the average gradient. We derive this optimal \( \beta \) (denoted \( \beta^* \)) using a second-order Taylor approximation.

\noindent
We begin with the natural-gradient update step of the form as  in Eq. \ref{eq:ngd_update}:
\begin{equation}
\theta_{\text{new}} = \theta - \eta F^{-1}(g_{\text{avg}} + \beta g_{\text{diff}}^\perp),
\label{eq:ng_step}
\end{equation}

\noindent
Using a second-order Taylor expansion of \( L_{\text{tot}} \) around \( \theta \), and assuming the Hessian matrix approximates to the Fisher matrix (\( \nabla^2 L_{\text{tot}} \approx F \)), the approximate loss after update is:
\begin{equation}
L_{\text{tot}}(\theta - \eta d) \approx L_{\text{tot}}(\theta) - \eta (g_1 + g_2)^\top d + \frac{\eta^2}{2} d^\top F d,
\label{eq:taylor_loss}
\end{equation}
where
\begin{equation}
d = F^{-1}(g_{\text{avg}} + \beta g_{\text{diff}}^\perp).
\label{eq:update_direction}
\end{equation}

\noindent
To isolate the effect of \( \beta \), we define a surrogate objective \( J_{\text{tot}}(\beta) \) by dropping constants and factors of \( \eta \):
\begin{align}
J_{\text{tot}}(\beta) = 
&- (g_1 + g_2)^\top F^{-1}(g_{\text{avg}} + \beta g_{\text{diff}}^\perp) \nonumber \\
&+ \frac{1}{2} (g_{\text{avg}} + \beta g_{\text{diff}}^\perp)^\top F^{-1}(g_{\text{avg}} + \beta g_{\text{diff}}^\perp).
\label{eq:surrogate_obj}
\end{align}

\noindent
To simplify, we define the following inner products:
\begin{equation}
D = g_{\text{avg}}^\top F^{-1} g_{\text{diff}}^\perp, \quad
E = (g_{\text{diff}}^\perp)^\top F^{-1} g_{\text{diff}}^\perp.
\label{eq:DE_defs}
\end{equation}

\noindent
Noting that \( g_1 + g_2 = 2 g_{\text{avg}} \), we substitute into \eqref{eq:surrogate_obj} and expand:
\begin{align}
J_{\text{tot}}(\beta) 
&= -2 D - 2\beta D + \frac{1}{2} \left( \|g_{\text{avg}}\|^2_{F^{-1}} + 2\beta D + \beta^2 E \right) \nonumber \\
&= -2D - \beta D + \frac{1}{2} \beta^2 E + \text{const},
\label{eq:expanded_J}
\end{align}
where \( \|g_{\text{avg}}\|^2_{F^{-1}} = g_{\text{avg}}^\top F^{-1} g_{\text{avg}} \) is absorbed into the constant term.

\noindent
To find the optimal \(\beta^*\) , we differentiate \eqref{eq:expanded_J} with respect to \( \beta \) and set the derivative to zero:
\begin{equation}
\frac{d J_{\text{tot}}}{d \beta} = -D + \beta^* E = 0 \quad \Rightarrow \quad \beta^* = \frac{D}{E}.
\label{eq:beta_opt}
\end{equation}

\noindent
Substituting back the definitions of \( D \) and \( E \), we obtain the closed-form expression:
\begin{equation}
\beta^* = \frac{g_{\text{avg}}^\top F^{-1} g_{\text{diff}}^\perp}
{(g_{\text{diff}}^\perp)^\top F^{-1} g_{\text{diff}}^\perp}.
\label{eq:beta_opt_explicit}
\end{equation}


\noindent
This yields a layer‐wise coefficient \(\beta^*\) that minimizes our second‐order surrogate, injecting orthogonal corrections only when beneficial. While it relies on the Hessian–Fisher approximation, which can misestimate curvature early in training or in highly nonlinear models, damped Fisher matrices and large‐batch averaging curb these errors. In practice, any misleading orthogonal signal drives \(\beta^*\to0\), safely reducing FOP to a standard KFAC update.

\subsection{Layer-wise Adaptive Scaling Step Size \( \eta_\ell^* \)}

\noindent
To ensure that each layer's update magnitude is automatically adjusted to account for its local curvature and gradient alignment, we introduce a layer-wise adaptive coefficient \( \eta_\ell^* \). Rather than using a single global learning rate for all parameters, \( \eta_\ell^* \) is chosen to (locally) minimize a quadratic approximation of the loss function along the natural-gradient direction for each layer \( \ell \).
Instead of differentiating about \(\beta\) in Eq. \ref{eq:taylor_loss}, 
we minimize this one-dimensional quadratic model in \( \eta \). After we set the derivative with respect to \( \eta \) to zero, the optimal step size becomes:
\begin{equation}
\eta_\ell^* = 
\frac{g_{\ell,\text{tot}}^\top F_\ell^{-1} g_{\ell,\text{comb}}}
{g_{\ell,\text{comb}}^\top F_\ell^{-1} g_{\ell,\text{comb}}}.
\label{eq:eta_l_opt}
\end{equation}

\noindent
This expression automatically adjusts the step size based on both the alignment between the total gradient and the proposed update direction, and the curvature in that direction. 
\noindent
When the curvature along \( g_{\ell,\text{comb}} \) is low and the update is well-aligned with the descent direction, \( \eta_\ell^* \) will approach 1, allowing a full natural-gradient step. Conversely, when the curvature is high or the combined gradient is poorly aligned with the total gradient, \( \eta_\ell^* \) decreases below 1, effectively damping the update to avoid overshooting.
The final updates are:

\begin{equation}
d_\ell = \eta_0 \eta_\ell^* F_\ell^{-1} g_{\ell,\text{comb}},
\label{eq:d_l_eta}
\end{equation}
where \(\eta_0\) is the base learning rate shared across the whole model.

\subsection{Kullback–Leibler (KL) norm analysis}

Natural-gradient methods, such as KFAC~\cite{martens2015optimizing}, select update directions that maximize progress in parameter space relative to the KL-divergence between the model before and after the update. A standard second-order approximation of the KL-divergence gives rise to the KL-norm:
\begin{equation}
\mathrm{KL}(p_{\theta + \Delta} \| p_\theta) \approx \frac{1}{2} \|F^{1/2} \Delta\|^2
\end{equation}
where \( F \) is the Fisher information matrix and \( \Delta \) is the parameter update step. The KL-norm quantifies how much the model distribution changes due to an update.

\noindent
In our case, the FOP update step is defined as:
\begin{equation}
\Delta_{\mathrm{FOP}} = -\eta M^{-1}(g + \beta g^\perp)
\end{equation}

where \( g \) is the average micro-batch gradient \( g_\text{avg} \), \( g^\perp \) is an orthogonal correction term in Eq. \ref{eq:combined_gradient}, satisfying \( (g^\perp)^\top F g = 0 \), \( M = F + \lambda I \) is the damped Fisher matrix, and \( \beta \in \mathbb{R} \) controls the contribution of the orthogonal component. 

\noindent
Substituting this into the KL-norm expression, we obtain a decomposition into three terms:

\begin{align}
\label{eq:KL-norm}
\left\| F^{1/2} \Delta_{\text{FOP}} \right\|^2 
&=\eta^2 \left[ 
g^\top Q g 
+ 2\beta g^\top Q g^\perp 
+ \beta^2 (g^\perp)^\top Q g^\perp 
\right]
\end{align}

\begin{itemize}
    \item A base-term \( g^\top Q g \) corresponding to standard KFAC,
    \item A cross-term \( 2\beta\, g^\top Q g^\perp \),
    \item And an orthogonal-term \( \beta^2 (g^\perp)^\top Q g^\perp \),
\end{itemize}

\noindent
where
\(
Q = (F + \lambda I)^{-1} F (F + \lambda I)^{-1}
\) acts as a curvature–weighted metric. In the large-damping case, where \( \lambda \gg \Lambda_i \) for every Fisher eigenvalue \( \Lambda_i \) that carries weight in \( g \), the spectral factor
\(
\frac{\Lambda_i}{(\Lambda_i + \lambda)^2}
\)
is proportional to \( \Lambda_i / \lambda^2 \). Therefore, the base term scales as
\(
g^\top Q g \propto \lambda^{-2}.
\)
\noindent
As shown in Supplementary material in the extended version, the two additional terms from FOP decay even more gently. Specifically:
\begin{align}
\label{eq:final kl norm}
\|F^{1/2} \Delta_{\text{FOP}}\|^2 
&\leq \eta^2 \Bigg[
\underbrace{g^\top Q g}_{\|F^{1/2} \Delta_{\text{KFAC}}\|^2} \nonumber \\
&+ 2\beta \frac{\mu_{\max}}{\lambda} \|F^{-1/2} g\| \cdot \|F^{-1/2} g_\perp\| \notag \nonumber \\
&\quad + \beta^2 \frac{\mu_{\max}}{4\lambda} \|F^{-1/2} g_\perp\|^2
\Bigg]
\end{align}




\noindent
Because the KL-norm of the FOP update splits into a base-term that decays as $\mathcal{O}(1/\lambda^2)$ and two correction terms that decay only as $\mathcal{O}(1/\lambda)$, there is an inherent separation in how quickly these components diminish as the damping parameter $\lambda$ increases. 

\noindent
During the early phase of training, it is common for the$\|g_{\text{avg}}\|$ to be large, while the$\|g_{\text{diff}^\perp}\|$ is also notable and dominated by high-frequency, low-curvature noise that is exaggerated by the application of $F^{-1}$. In such scenarios, the orthogonal correction $g^\perp_{\text{diff}}$ often points in the opposite direction to the main descent path. As a result, the optimal mixing coefficient becomes negative ($\beta < 0$), leading to a negative cross-term in the KL-norm. This partial cancellation of the core component creates margins that safely reduce the damping factor $\lambda$.

\subsection{Distributed FOP}

\begin{algorithm}[h]
\caption{Distributed FOP with Dual\,Gradients}
\label{alg:proj_kfac_concise}
\begin{algorithmic}[1]
\Require
  \Statex $M$ : neural network model
  \Statex $P=\{p_0,\dots,p_{N-1}\}$ : set of $N$ GPU processes
  \Statex $S_j$ : subset of layers for which $p_j$ is the curvature \emph{specialist}
  \Statex $\mathcal{D}_j$ : local mini‐batch on $p_j$
  \Statex $F_i$ : running estimate of the Fisher matrix block for layer $\ell_i$,
          stored and updated by its specialist GPU $p_k$
\Statex
  \Statex $G_{\text{pri}}=\{p_j\in P \mid j \bmod 2 = 0\}$  \Comment{primary group}
  \Statex $G_{\text{sec}}=\{p_j\in P \mid j \bmod 2 = 1\}$  \Comment{secondary group}
\Ensure preconditioned global gradient $\tilde g$

\ForAll{$p_j \in P$ \textbf{in parallel}}
  \State $g_j \gets \nabla_{\theta} \mathcal{L}(M;\mathcal{D}_j)$  \Comment{local back‑prop}
  \If{curvature update step}
    \ForAll{layers $\ell_i$ in $M$}
      \State $p_k \gets$ specialist s.t.\ $\ell_i\in S_k$
      \State send local curvature factors of $\ell_i$ to $p_k$
      \Statex \hfill /* async, non‑blocking */
      \If{$j=k$}
        \State update and invert $F_i \rightarrow F_i^{-1}$
        
      \EndIf
    \EndFor
  \EndIf

  \If{$p_j \in G_{\text{pri}}$}
    \State global \textsc{AllReduce} to compute $g_1$
  \Else
    \State global \textsc{AllReduce} to compute $g_2$
  \EndIf

  \ForAll{layers $\ell_i$ in $M$}
    \State $p_k \gets$ specialist for $\ell_i$
    \If{$j = k$}
      \State $\tilde g_i \gets \textsc{FOP}(g_{1,i}, g_{2,i},  F_i^{-1})$
      \State broadcast $\tilde g_i$ to all processes
    \EndIf
  \EndFor
\EndFor
\State assemble $\tilde g \gets [\tilde g_1,\dots,\tilde g_L]$  \Comment{$L$ = \#layers}
\end{algorithmic}
\end{algorithm}

\noindent
Traditional second-order optimization methods like KFAC are notoriously difficult to scale due to the high memory and computation costs of storing and inverting large curvature matrices. Moreover, synchronizing these matrices across multiple GPUs introduces significant communication overhead, making them impractical for large-scale training without careful system design.
We design a scalable FOP implementation that combines data parallelism with lightweight model parallelism to minimize the overhead of splitting large batches. First, we assign each GPU as a specialist, responsible for updating the curvature (Fisher information) of a subset of layers via a sharded preconditioner similar to the previous works~\cite{osawa2023asdl,pauloski2022deep}. 
Second, we introduce a dual-gradient reduction strategy, where two global gradients \( \mathbf{g}_1 \) and \( \mathbf{g}_2 \) are computed in parallel over disjoint GPU groups.
Finally, each specialist GPU applies the FOP using its local Fisher inverse and both gradients to compute its layer's update, which is then broadcast across the GPUs. 
This distributed design enables efficient second-order updates across large-scale multi-GPU systems. The full algorithm is shown in Algorithm~\ref{alg:proj_kfac_concise}.

\section{Experiments}
\begin{figure}
    \centering
    \includegraphics[width=0.9\linewidth]{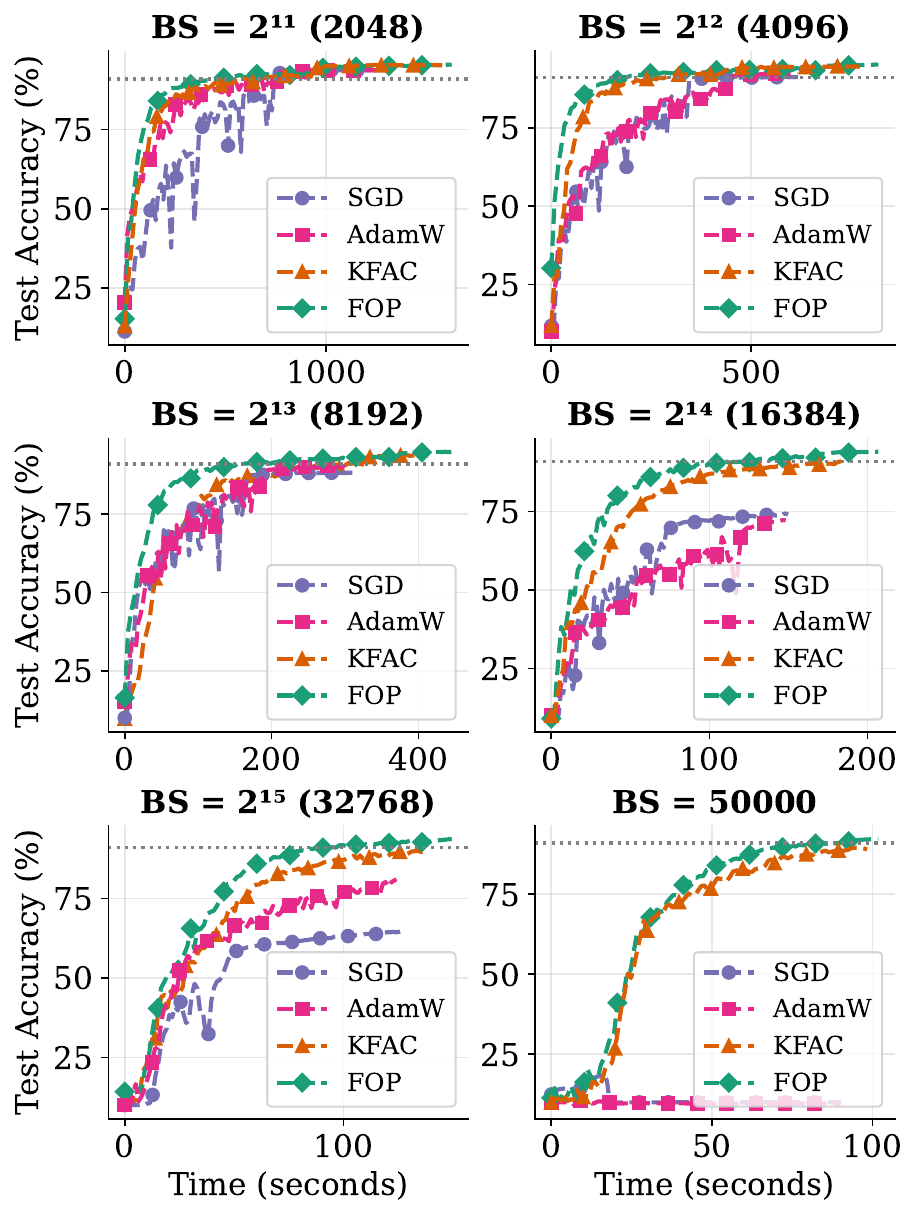}
    \caption{
        Test accuracy vs. wall-clock time (in seconds) for ResNet-18 on CIFAR-10, grouped by batch size. The dotted line represents the threshold of 91\%. 
    }
    \label{fig:cifar10_batch_time}
\end{figure}

\noindent In this section, we rigorously evaluate FOP against both first‐order (SGD, AdamW) and second‐order (KFAC) baselines across four vision benchmarks: CIFAR-10 with ResNet-18, ImageNet-100 with T2T-ViT, ImageNet-1K with ResNet-50, and long-tailed CIFAR10-LT/100-LT with ResNet-32, demonstrating its fast convergence, large-batch scalability, and robustness under class imbalance. 
\noindent
To ensure fair and rigorous comparisons, we evaluate all methods on several standard benchmarks: ResNet-18 on CIFAR-10, ResNet-50 on ImageNet-1k, and a Vision Transformer (ViT) on ImageNet-100. For each setting, we perform an extensive hyperparameter search for all optimizers. This includes tuning the learning rate, and for second-order methods like KFAC and FOP, also tuning the damping ratio \( \lambda \). Following the linear scaling rule from \cite{goyal2017accurate}, the learning rate is scaled proportionally with the batch size. Additionally, the curvature update frequency for the second-order optimizer is reduced as the batch size increases, until it reaches a minimum threshold of 5 steps. To isolate the effect of our Fisher-orthogonal projection, FOP and KFAC share identical learning-rate schedules in every experiment. 
\noindent
While our results highlight FOP's advantages, we do not claim that FOP is a universally superior optimizer for all tasks and architectures. Rather, the evidence shows that augmenting natural gradient methods with a principled, geometry-aware variance component offers via FOP a robust and scalable path for second-order optimization in modern large-batch training scenarios.
\noindent
All experiments were performed on a single node with two AMD EPYC 9534 64-core CPUs and eight AMD MI300X GPUs. The implementations of KFAC and FOP rely on the ASDL package ~\cite{osawa2023asdl}.
Full details of the hyperparameter search space, such as learning rate, damping rate, and random seeding number, and a discussion about the memory overhead are provided in the Supplementary material in the extended version.

\subsection{CIFAR10 with ResNet18}

\noindent
We first evaluate optimization performance on the CIFAR-10 dataset~\cite{cifar}, a widely used benchmark for assessing second-order optimizers~\cite{eschenhagen2023kronecker,liu2024layer,martens2015optimizing}. Each experiment is run with 5 different random seeds to ensure robustness. We employ the \texttt{ReduceLROnPlateau} learning rate scheduler during training. Figure~\ref{fig:cifar10_batch_time} illustrates the progression of test accuracy over wall-clock time for ResNet-18, across batch sizes ranging from $2048$ to $50000$ (the total number of training samples in CIFAR-10). At small to moderate batch sizes (e.g., $2048$ and $4096$), all optimizers, including SGD, AdamW, KFAC, and FOP, achieve 91\% accuracy, though FOP consistently reaches this threshold the fastest. As batch size increases beyond $16384$, first-order optimizers such as SGD and AdamW struggle to converge within the same epoch limit, failing to reach 91\% accuracy altogether at larger batch sizes (e.g., $32768$ and $50000$).

\noindent
These trends are quantitatively summarized in Table~\ref{tab:imagenet91}, which reports both the epochs and total wall‐clock time to reach 91\% Top-1 accuracy on CIFAR-10 using the same GPU count. At BS=2048, FOP hits the target accuracy in 29/475.2s, with $\times1.56$ faster than SGD (58/743.3s) and $\times1.26$ faster than KFAC (37/588.7s). As we scale to BS=4096 and 8192, FOP’s speedup over SGD grows to $\times1.69$ and $\times2.91$ , respectively, and reaches $\times3.78$ at BS=16384. Crucially, FOP is the only method to arrive 91\% at BS=32768 and 50000, doing so in 60/90.6s ($\times5.05$) and 82/84.3s ($\times5.43$). These results underscore FOP’s exceptional large-batch scalability and its ability to deliver substantial accelerations.

\begin{table}
\centering
\renewcommand{\arraystretch}{0.8}
\small
\begin{tabular}{@{}c|cccc@{}}
\toprule
\textbf{Batch Size}
  & \multicolumn{4}{c}{\textbf{Epochs / Time (s) and Speedup}} \\ 
  \cmidrule{2-5}
\textbf{(GPU)}  & \textbf{SGD} & \textbf{AdamW} & \textbf{KFAC} & \textbf{FOP (ours)} \\
\midrule
2048 (2)
  & 58  / 743.3   & 61  / 768.4   
  & 37  / 588.7   
  & 29  / 475.2    \\
\addlinespace
\multirow{2}{*}{4096 (2)}  
  & 73  / 457.9   & 73  / 454.0   
  & 34  / 270.5   
  & 22  / 181.9    \\
  & –            & –            
  & [$\times1.69$]  
  & [$\times2.52$]   \\
\addlinespace
\multirow{2}{*}{8192 (2)}  
  & -- / --       & -- / --      
  & 71  / 296.4   
  & 35  / 157.5    \\
  & --           & --           
  & [$\times1.54$]  
  & [$\times2.91$]   \\
\addlinespace
\multirow{2}{*}{16384 (2)} 
  & -- / --       & -- / --      
  & 99  / 186.4   
  & 58  / 121.2    \\
  & --           & --           
  & [$\times2.46$]  
  & [$\times3.78$]  \\
\addlinespace
\multirow{2}{*}{32768 (2)} 
  & -- / --       & -- / --      
  & -- / --       
  & 60  / 90.6     \\
  & --           & --           
  & --           
  & [$\times5.05$]   \\
\addlinespace
\multirow{2}{*}{50000 (2)} 
  & -- / --       & -- / --      
  & -- / --       
  & 82  / 84.3     \\
  & --           & --           
  & --           
  & [$\times5.43$]   \\
\bottomrule
\end{tabular}
\caption{
Epoch and training time (in seconds) to reach 91\% Top-1 accuracy on CIFAR-10. Each row shows the batch size and number of GPUs used in the format: \emph{Batch (GPU)} and the corresponding \emph{Epoch/Time}. “--” indicates the accuracy threshold was not reached. For KFAC and FOP, the bracketed numbers show the speedup factor relative to SGD at the batch size of 4096.
}
\label{tab:imagenet91}
\end{table}

\begin{figure}[h]
    \centering
    \includegraphics[width=\linewidth]{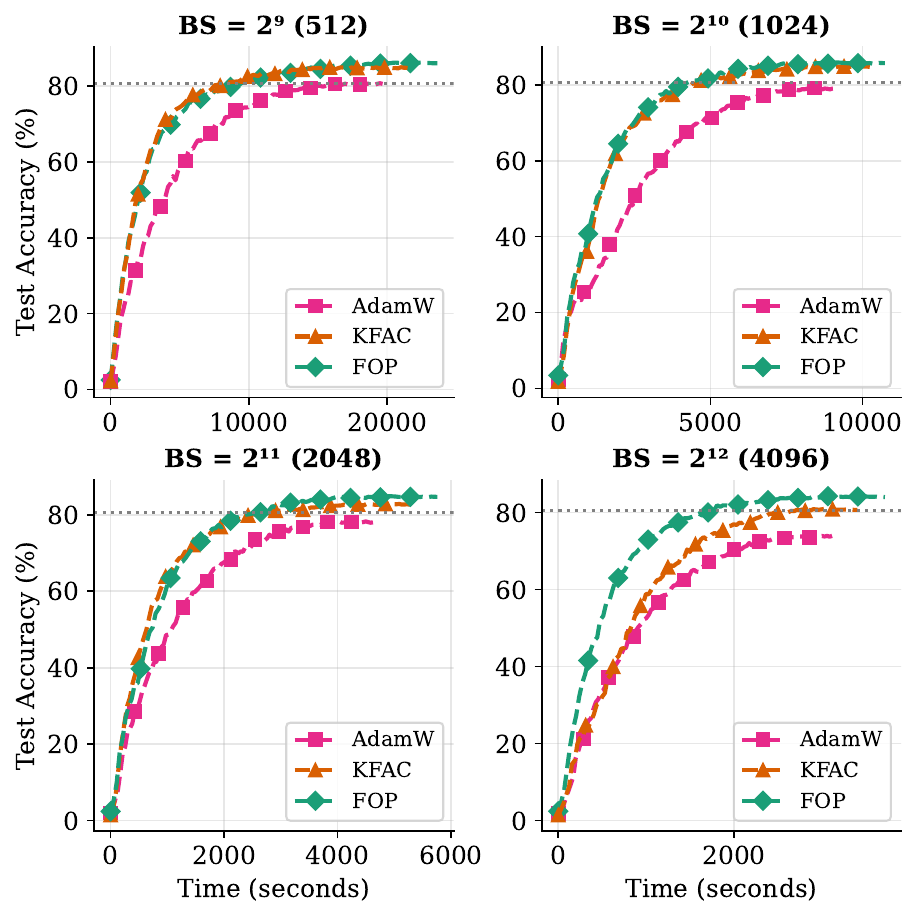}
    \caption{
        Test accuracy vs. wall-clock time (in seconds) for T2T-ViT on ImageNet-100, grouped by batch size. The dotted line represents the threshold of 80.6\%. 
    }
    \label{fig:imagenet100_batch_time}
\end{figure}

\subsection{ImageNet-100 with T2T-ViT}

\noindent
To evaluate FOP on modern transformers, we train a Tokens-to-Token Vision Transformer (T2T-ViT) from scratch on ImageNet-100 with running 3 different random seeds\citep{Yuan_2021_ICCV,lin2024structured}. Following \citet{lin2024structured}, we apply FOP and KFAC only to the convolutional and linear layers’ gradients and activations, while all other parameters (e.g., LayerNorm) are updated with AdamW \citep{adamw}. We run 100 epochs with a cosine learning‐rate schedule and measure Top-1 accuracy over wall-clock time (Figure~\ref{fig:imagenet100_batch_time}). We set our target at 80.6\%, the best result achieved by AdamW at batch size 512, and report the epochs and training time to reach it in Table~\ref{tab:vit_batchsize_80_results}.

\noindent
AdamW requires nearly the full 100 epochs, and the longest runtime at batch size 512, whereas both second-order methods hit 80.6\% in substantially fewer epochs and less time. FOP consistently outperforms KFAC and AdamW across batch sizes larger than 512, delivering speedups of $\times4.33$, $\times6.90$, and $\times10.48$, where KFAC only achieves $\times3.80$, $\times6.34$, and $\times6.45$ compared to AdamW. KFAC scales better than AdamW but still lags behind FOP, typically needing more epochs to match the same accuracy.

\begin{table}[h]
\centering
\renewcommand{\arraystretch}{0.8}
\small

\begin{tabular}{@{}l|cccc@{}}
\toprule
\multirow{2}{*}{\textbf{Optimizer}} &
  \multicolumn{4}{c}{\textbf{Batch Size (GPU): Epochs / Time (min)}} \\
\cmidrule(lr){2-5}
 & \textbf{512 (1)} & \textbf{1024 (2)} & \textbf{2048 (4)} & \textbf{4096 (8)} \\
\midrule
\multirow{2}{*}{AdamW}
  & 97 / \num{291.6} & --             & --             & --              \\
  &                  &                &                &                 \\

\multirow{2}{*}{KFAC}
  & 42 / \num{138.6} & 49 / \num{76.7} & 57 / \num{46.0} & 87 / \num{45.3} \\
  & [\(\times2.10\)] & [\(\times3.80\)] & [\(\times6.34\)] & [\(\times6.45\)] \\

\multirow{2}{*}{FOP (ours)}
  & 44 / \num{158.9} & 41 / \num{67.3} & 48 / \num{42.3} & 49 / \num{27.8} \\
  & [\(\times1.84\)] & [\(\times4.33\)] & [\(\times6.90\)] & [\(\times10.48\)] \\
\bottomrule
\end{tabular}

\caption{
Epoch and training time (in minutes) to reach 80.6\% Top-1 accuracy for T2T-ViT on ImageNet-100. 
For KFAC and FOP, the bracketed numbers show the speedup factor relative to AdamW at the batch size of 512.}
\label{tab:vit_batchsize_80_results}
\end{table}




\subsection{ImageNet-1K with ResNet50}

\begin{figure}
    \centering
    \includegraphics[width=\linewidth]{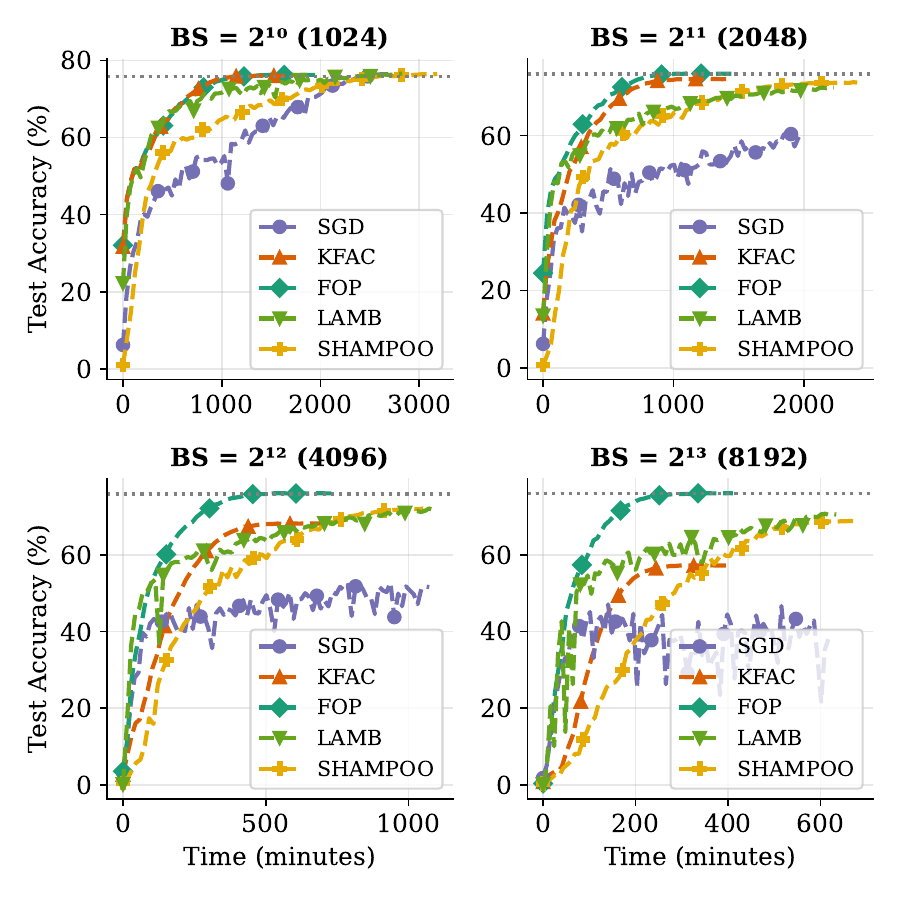}
    \caption{
        Test accuracy vs. wall-clock time (in minutes) for ResNet-50 on ImageNet-1K, grouped by batch size. The dotted line represents the threshold of 75.9\%.
    }
    \label{fig:imagenet1k_batch_time}
\end{figure}

\begin{table}[h]
\centering
\renewcommand{\arraystretch}{0.8}
\small
\begin{tabular}{@{}l|cccc@{}}
\toprule
\multirow{2}{*}{\textbf{Optimizer}} &
  \multicolumn{4}{c}{\textbf{Batch Size (GPU): Epochs / Time (min)}} \\
\cmidrule(lr){2-5}
 & \textbf{1024 (1)} & \textbf{2048 (2)} & \textbf{4096 (4)} & \textbf{8192 (8)} \\
\midrule
\multirow{2}{*}{SGD}
  & 71 / 2511.1 & -- / --    & -- / --    & -- / --    \\
  &             &            &            &            \\

\multirow{2}{*}{Shampoo}
  & 65 / 2523.1 & -- / --    & -- / --    & -- / --    \\
  &             &            &            &            \\

\multirow{2}{*}{LAMB}
  & 67 / 2492.7 & -- / --    & -- / --    & -- / --    \\
  &             &            &            &            \\

\multirow{2}{*}{KFAC}
  & 35 / 1336.5 & -- / --    & -- / --    & -- / --    \\
  & [(\(\times1.88\))] &     &            &            \\

\multirow{2}{*}{FOP (ours)}
  & 32 / 1306.1 & 33 /  999.5 & 34 /  514.9 & 40 /  335.1 \\
  & [(\(\times1.92\))] & [(\(\times2.51\))] & [(\(\times4.88\))] & [(\(\times7.50\))] \\
\bottomrule
\end{tabular}

\caption{
Epoch and training time (in minutes) to reach 75.9\% Top-1 accuracy on ImageNet-1K with ResNet-50. 
For KFAC and FOP, the bracketed numbers show the speedup factor relative to SGD at the batch size of 1024.
}
\label{tab:imagenet1k_batchsize_80_results}
\end{table}


\noindent
To evaluate FOP at a much larger-scale dataset, we train ResNet-50 from scratch on ImageNet-1K \citep{deng2009imagenet} with running 3 different random seeds, to see if it can reach Top-1 test accuracy of 75.9\%, which is a standard large-scale convolutional benchmark \cite{mattson2019mlperf}. Following \citet{yang2020sketchy,liu2024layer}, we run SGD, Shampoo \cite{gupta2018shampoo}, and LAMB \cite{you2019lamb} for 80 epochs with a cosine learning-rate schedule. We train both KFAC and FOP for 50 epochs using an exponential schedule on their learning rates, because in the SGD, Shampoo, and LAMB can't reach the threshold of 75.9\% within 50 epochs.

\noindent
Figure~\ref{fig:imagenet1k_batch_time} and Table~\ref{tab:imagenet1k_batchsize_80_results} illustrate the dramatic efficiency gains of FOP in reaching 75.9\% Top‐1 accuracy on ImageNet‐1K. At BS=1024, FOP converges in 32/1306.1 min, that is nearly $\times2$ faster than SGD (71/2511.1 min) and just slightly ahead of K‐FAC’s 35/1336.5 min ($\times1.88$). Beyond this scale, only FOP succeeds to hit the threshold in 33/999.5 min at BS=2048 ($\times2.51$), 34/514.9 min at BS=4096 ($\times4.88$), and 40/335.1 min at BS=8192 ($\times7.50$), while both SGD and K‐FAC stall below 75.9\%. FOP’s steadily shrinking time‐to‐threshold with increasing batch size demonstrates its better large‐batch scalability and clear advantage over first‐ and second‐order alternatives. 
Moreover, our FOP results compare favorably even to recent second‐order methods. For instance, SENG \cite{yang2020sketchy} reaches 75.9\% Top‐1 on ImageNet‐1K in 41 epochs at BS=4096, and SP-NGD \cite{osawa2020scalable} only hits 74.8\%–75.3\% at BS=4096–8192. In contrast, FOP matches or exceeds their results in fewer epochs and achieves the same threshold in 34 epochs at BS=4096 and 40 epochs at BS=8192.

\subsection{CIFAR10-LT with ResNet32}

Finally, to assess optimizer robustness under long‐tailed data, we evaluate KFAC and FOP on the CIFAR10‐LT and CIFAR100‐LT benchmarks (imbalance factors of 100 and 50) using a lightweight ResNet-32, following the training protocol of \citet{zhang2021tricks}. We apply both second‐order methods as preconditioners with a fixed damping ratio of $1\mathrm{e}{-3}$ and obtain results over 5 random seeds.

\noindent
Table~\ref{tab:cifar_lt_accuracy} reports the Top-1 error rates on the CIFAR‐LT datasets under two imbalance factors (50 and 100). We compare our implementations of KFAC and FOP against baseline results from~\citet{zhang2021tricks} and representative results from other recent works~\cite{liu2019large,kang2019decoupling,cui2019class}. FOP consistently delivers the lowest error, outperforming the baseline by 2.3–3.3\%  smaller error rate and KFAC by 1.8–2.0\% smaller error rate across all settings. For example, on CIFAR-100-LT with factor 100, it cuts the error from 62.27\% (baseline) to 58.97\% (3.3\% drop), and on CIFAR-10-LT with factor 50, it reduces error from 23.55\% to 20.55\% (3.0\% drop), highlighting its robustness under severe class imbalance. By contrast, KFAC delivers only modest reduction ($\approx 1\%$) on the CIFAR-LT dataset with an imbalance factor of 50, and even underperforms the baseline/other works on those with the factor of 100 (28.59\% vs. 28.05\% in CIFAR-10-LT, and 61.78\% vs. 61.68\% in CIFAR-100-LT). These results highlight FOP’s superior robustness under severe class imbalance, a benefit we attribute to the Fisher–Orthogonal Projection term that balances curvature estimates across head and tail classes.

\begin{table}[h]
\centering
\renewcommand{\arraystretch}{0.9}
\small
\begin{tabular}{@{}lcccc@{}}
\toprule
\textbf{Datasets} & \multicolumn{2}{c}{CIFAR-10-LT} & \multicolumn{2}{c}{CIFAR-100-LT} \\
\cmidrule(lr){2-3} \cmidrule(lr){4-5}
\textbf{Imbalance Factor} & 100 & 50 & 100 & 50 \\
\textbf{Backbones} & \multicolumn{4}{c}{ResNet-32} \\
\midrule
Baselines   & 28.05 & 23.55 & 62.27 & 56.22 \\
Other works & 29.64 & 25.19 & 61.68 & 56.15 \\
KFAC        & 28.59 \(\uparrow\) & 22.29 \(\downarrow\)  & 61.78 \(\uparrow\) & 55.02 \(\downarrow\)  \\
FOP (ours)  & \textbf{26.65} \(\downarrow\) & \textbf{20.55} \(\downarrow\) & \textbf{58.97} \(\downarrow\) & \textbf{53.67} \(\downarrow\) \\
\bottomrule
\end{tabular}
\caption{Top-1 error rates on CIFAR-LT, comparing KFAC and FOP against the implementation baseline~\cite{zhang2021tricks} and prior results~\cite{liu2019large,kang2019decoupling,cui2019class}.
\(\downarrow\) indicates that an error rate is lower (better) than both the baseline and other reported results;  
\(\uparrow\) indicates that it is higher (worse).
}
\label{tab:cifar_lt_accuracy}
\end{table}

\section{Conclusion}

In this work, we introduced Fisher–Orthogonal Projection (FOP), a novel second-order optimizer that combines natural-gradient updates with a geometry-aware variance correction. This design eliminates the need for task-specific tuning when scaling to multiple GPUs in large-batch training. FOP enables seamless scaling to extremely large batch sizes, without requiring any additional tricks or adaptive hyperparameter adjustments, except scaling the learning rate as the batch size. In contrast, standard optimizers such as SGD, AdamW, and KFAC often collapse or demand extensive tuning under such conditions.
We validate FOP through extensive experiments on four diverse benchmarks: ResNet-18 on CIFAR-10, T2T-ViT on ImageNet-100, ResNet-50 on ImageNet-1K, and ResNet-32 on long-tailed CIFAR-LT. Across these settings, FOP consistently accelerates convergence and scales more robustly to large batches. Moreover, under severe class imbalance in CIFAR-LT dataset, FOP delivers better generalization, reducing Top-1 error by 2.3–3.3\% compared to strong baselines.
Together, these results highlight FOP’s unique ability to unlock stable, plug-and-play large-batch training across both convolutional and transformer architectures. This makes it especially well-suited for large-scale distributed and resource-constrained environments, paving the way for practical, reliable second-order optimization at scale.

\section{Acknowledgments}
This work was supported by UK Research and Innovation (UKRI) EP/T022205/1. The authors would also like to acknowledge support from the JADE 2.5 supercomputing service. The authors would also like to acknowledge the use of the University of Oxford Advanced Research Computing (ARC) facility \citep{richards2015universityarc}.
The authors also acknowledge the use of resources provided by the Isambard-AI National AI Research Resource (AIRR). Isambard-AI is operated by the University of Bristol and is funded by the UK Government’s Department for Science, Innovation and Technology (DSIT) via UK Research and Innovation; and the Science and Technology Facilities Council [ST/AIRR/I-A-I/1023].





\begin{center}
    {\LARGE \textbf{Supplementary Material}}\\[1em]
\end{center}
\vspace{1em}

\section{A.  Notations and Lemmas}

\begin{lemma}
    
\label{lem:lemma1}

Let \( g_{\text{diff}}, g_{\text{avg}} \in \mathbb{R}^n \), and let \( F \in \mathbb{R}^{n \times n} \) be a symmetric positive semi-definite matrix (the Fisher information matrix). Define the scalar projection:
\[
s_{\text{FOP}} := \frac{g_{\text{diff}}^\top F g_{\text{avg}}}{g_{\text{avg}}^\top F g_{\text{avg}} + \epsilon}
\]
and the orthogonal component:
\[
g_{\text{diff}}^\perp := g_{\text{diff}} - s_{\text{FOP}} \, g_{\text{avg}}
\]
for some small constant \( \epsilon \). Then the Fisher inner product between \( g_{\text{diff}}^\perp \) and \( g_{\text{avg}} \) satisfies:
\[
\left( g_{\text{diff}}^\perp \right)^\top F g_{\text{avg}} = 
g_{\text{diff}}^\top F g_{\text{avg}} \cdot 
\left( \frac{\epsilon}{g_{\text{avg}}^\top F g_{\text{avg}} + \epsilon} \right)
\]
In particular:
\begin{itemize}
    \item If \( \epsilon = 0 \), then \( g_{\text{diff}}^\perp \) is exactly \( F \)-orthogonal to \( g_{\text{avg}} \).
    \item For \( \epsilon > 0 \), the deviation from orthogonality is bounded by:
    \[
    \left| \left( g_{\text{diff}}^\perp \right)^\top F g_{\text{avg}} \right|
    \leq \epsilon \, \|F^{1/2} g_{\text{diff}}\| \cdot \|F^{-1/2} g_{\text{avg}}\|
    \]
\end{itemize}

\begin{proof}

We start with the definition:
\[
g_{\text{diff}}^\perp = g_{\text{diff}} - s_{\text{FOP}} \, g_{\text{avg}},
\quad
s_{\text{FOP}} = \frac{g_{\text{diff}}^\top F g_{\text{avg}}}{g_{\text{avg}}^\top F g_{\text{avg}} + \epsilon}
\]

\noindent Now take the Fisher inner product of \( g_{\text{diff}}^\perp \) with \( g_{\text{avg}} \):
\begin{align*}
\left( g_{\text{diff}}^\perp \right)^\top F g_{\text{avg}} 
&= g_{\text{diff}}^\top F g_{\text{avg}} - s_{\text{FOP}} \cdot g_{\text{avg}}^\top F g_{\text{avg}} \\
&= g_{\text{diff}}^\top F g_{\text{avg}} - 
\frac{g_{\text{diff}}^\top F g_{\text{avg}}}{g_{\text{avg}}^\top F g_{\text{avg}} + \epsilon} \cdot g_{\text{avg}}^\top F g_{\text{avg}} \\
&= g_{\text{diff}}^\top F g_{\text{avg}} \left( 
1 - \frac{g_{\text{avg}}^\top F g_{\text{avg}}}{g_{\text{avg}}^\top F g_{\text{avg}} + \epsilon} 
\right) \\
&= g_{\text{diff}}^\top F g_{\text{avg}} \cdot 
\frac{\epsilon}{g_{\text{avg}}^\top F g_{\text{avg}} + \epsilon}
\end{align*}

\noindent This proves the first part of the lemma. If \( \epsilon = 0 \), the expression is zero, yielding exact orthogonality:
\[
\left( g_{\text{diff}}^\perp \right)^\top F g_{\text{avg}} = 0
\]

\noindent For \( \epsilon > 0 \), we apply the Cauchy-Schwarz inequality under the Fisher inner product:
\begin{align*}
\left| \left( g_{\text{diff}}^\perp \right)^\top F g_{\text{avg}} \right|
&= \left| g_{\text{diff}}^\top F g_{\text{avg}} \cdot 
\frac{\epsilon}{g_{\text{avg}}^\top F g_{\text{avg}} + \epsilon} \right| \\
&\leq \epsilon \cdot \|F^{1/2} g_{\text{diff}}\| 
\cdot \|F^{-1/2} g_{\text{avg}}\|
\end{align*}
\end{proof}
\end{lemma}

\begin{lemma} 
\label{lem:lemma2}
Let \( F \succ 0 \) be a symmetric positive definite matrix, and let \( \mu_{\min} \) and \( \mu_{\max} \) denote the minimum and maximum eigenvalues of \( F \), respectively. Then for any vector \( x \in \mathbb{R}^n \), the following inequality holds:
\[
\mu_{\min}^{1/2} \, \|F^{-1/2} x\| \leq \|x\| \leq \mu_{\max}^{1/2} \, \|F^{-1/2} x\|.
\]
\end{lemma}

\begin{proof}
Since \( F \succ 0 \), it is diagonalizable with positive eigenvalues \( \mu_1, \dots, \mu_n \) and orthonormal eigenvectors. Let \( x \in \mathbb{R}^n \). The standard Euclidean norm is:
\[
\|x\|^2 = x^\top x,
\]
and the Fisher-scaled norm is:
\[
\|F^{-1/2} x\|^2 = x^\top F^{-1} x.
\]

\noindent From spectral theory, the matrix inequality holds:
\[
\mu_{\min} I \preceq F \preceq \mu_{\max} I,
\]
which implies (after inverting all sides):
\[
\mu_{\max}^{-1} I \preceq F^{-1} \preceq \mu_{\min}^{-1} I.
\]

\noindent Now apply these bounds to the quadratic form \( x^\top F^{-1} x \):
\begin{align*}
x^\top F^{-1} x &\leq \mu_{\min}^{-1} x^\top x = \mu_{\min}^{-1} \|x\|^2, \\
x^\top F^{-1} x &\geq \mu_{\max}^{-1} x^\top x = \mu_{\max}^{-1} \|x\|^2.
\end{align*}

\noindent Rewriting in terms of norms:
\begin{align*}
\|F^{-1/2} x\|^2 &\leq \mu_{\min}^{-1} \|x\|^2 
\quad \Rightarrow \quad \|x\| \geq \mu_{\min}^{1/2} \|F^{-1/2} x\|, \\
\|F^{-1/2} x\|^2 &\geq \mu_{\max}^{-1} \|x\|^2 
\quad \Rightarrow \quad \|x\| \leq \mu_{\max}^{1/2} \|F^{-1/2} x\|.
\end{align*}

\end{proof}

\section{B.  KL-norm analysis}
\subsection{B.1. KL-norm of FOP}
In natural gradient methods, such as K-FAC \cite{martens2015optimizing}, they define the
direction in parameter space which gives the largest change in the objective per unit of change in the model \cite{Amari2000MethodsOI}, as measured by the Kullback–Leibler (KL) divergence between the current model \( p_\theta \) and the updated model \( p_{\theta + \Delta} \).

By applying a second-order Taylor expansion of the KL divergence around \( \theta \), we obtain the following approximation:
\begin{equation}
\mathrm{KL}(p_{\theta + \Delta} \,\|\, p_\theta) 
\approx \frac{1}{2} \Delta^\top F \Delta 
= \frac{1}{2} \| F^{1/2} \Delta \|^2
\end{equation}
where \( F \) is the Fisher information matrix, and \( \| F^{1/2} \Delta \|^2 \) is known as the KL-norm of the update step \( \Delta \). This KL-norm measures how far the update moves the model in distribution space, making it a more meaningful metric for optimization in probabilistic models.

In our paper, the update rule in the Fisher Orthogonal Projection (FOP) method:
\begin{equation}
\Delta_{\mathrm{FOP}} = -\eta M^{-1} (g + \beta g^\perp)
\end{equation}
where:
\begin{itemize}
    \item \( M = F + \lambda I \) is the damped Fisher matrix,
    \item \( \beta \in \mathbb{R} \) controls the contribution from the orthogonal component \( g^\perp \),
    \item \( \eta \) is the learning rate,
    \item  \( g^\perp \equiv g_{\text{diff}}^\perp  \) and \( g \equiv g_{\text{avg}}  \)
    \item \( g^\perp   \) is orthogonal to \( g \) in the Fisher metric: \( (g^\perp)^\top F g = 0 \).
\end{itemize}

To compute the KL-norm of this FOP update, we evaluate:
\begin{equation}
    \|F^{1/2} \Delta_{\mathrm{FOP}} \|^2 
= \eta^2 (g + \beta g^\perp)^\top M^{-1} F M^{-1} (g + \beta g^\perp)
\end{equation}

Let
\begin{equation}
\label{eq:Q}
Q = M^{-1} F M^{-1}
\end{equation}

which is symmetric and positive semi-definite. Then the KL-norm becomes:

\begin{align}
\label{eq:KL-norm}
\left\| F^{1/2} \Delta_{\text{FOP}} \right\|^2 
&= \eta^2 (g + \beta g^\perp)^\top Q (g + \beta g^\perp) \nonumber \\
&=\eta^2 \left[ 
g^\top Q g 
+ 2\beta g^\top Q g^\perp 
+ \beta^2 (g^\perp)^\top Q g^\perp 
\right]
\end{align}

where \( g^\top Q g \) is the KL norm of KFAC, \( 2\beta \, g^\top Q g^\perp \) is a cross term and \( \beta^2 (g^\perp)^\top Q g^\perp \) is contributing the orthogonal projection. Each term captures a different component of the curvature-informed update. 

The first term \( g^\top Q g \) was proved to be situated in the trust region \cite{martens2015optimizing}, so we now focus on bounding the cross term \( g^\top Q g^\perp \), and the orthogonal term  \( \beta^2 (g^\perp)^\top Q g^\perp \). Bounding these terms is essential to ensure that the KL-norm remains predictable under varying damping parameters.

\subsubsection{Bounding \( 2\beta \, g^\top Q g^\perp \) \\}
\vspace{0.5em}

Substituting Eq. \ref{eq:Q} into the cross term gives:
\begin{align}
g^\top Q g^\perp 
&= g^\top M^{-1} g^\perp - \lambda g^\top M^{-2} g^\perp \label{eq:cross-split}
\end{align}
and we can bound this:

\begin{equation}
    |g^\top Q g^\perp| = \left| g^\top M^{-1} g^\perp - \lambda g^\top M^{-2} g^\perp \right|
\end{equation}
 and with triangle inequality (\(|a - b| \leq |a| + |b|\)):
 \begin{equation}
 \label{eq:triangle inequality}
     |g^\top Q g^\perp| \leq |g^\top M^{-1} g^\perp| + \lambda |g^\top M^{-2} g^\perp|
 \end{equation}
\noindent Applying the Cauchy–Schwarz inequality to each term:

\noindent For the first term:
\begin{equation}
|g^\top M^{-1} g^\perp| \leq \|M^{-1/2} g\| \cdot \|M^{-1/2} g^\perp\| \label{eq:cs-1}
\end{equation}

\noindent  For the second term:
\begin{equation}
|g^\top M^{-2} g^\perp| \leq \|M^{-1} g\| \cdot \|M^{-1} g^\perp\| \label{eq:cs-2}
\end{equation}

\noindent  Since \( M^{-1} \preceq \lambda^{-1} I \), it follows that:
\begin{align}
\|M^{-1/2} g\| &\leq \lambda^{-1/2} \|g\|, &
\|M^{-1} g\| &\leq \lambda^{-1} \|g\| \label{eq:M-bounds}
\end{align}

\noindent Putting Eq. \ref{eq:M-bounds} into Eq. \ref{eq:cs-1} and Eq. \ref{eq:cs-2}:
\begin{align}
\|M^{-1/2} g\| \cdot \|M^{-1/2} g^\perp\| 
&\leq \lambda^{-1} \|g\| \cdot \|g^\perp\| \\
\lambda \cdot \|M^{-1} g\| \cdot \|M^{-1} g^\perp\| 
&\leq \lambda \cdot \lambda^{-1} \cdot \lambda^{-1} \|g\| \cdot \|g^\perp\| \nonumber \\
\qquad & = \lambda^{-1} \|g\| \cdot \|g^\perp\|
\end{align}

\noindent Putting these back to Eq. \ref{eq:triangle inequality}, we get:
\begin{equation}
    |g^\top Q g^\perp| \leq \lambda^{-1} \|g\| \cdot \|g^\perp\| + \lambda^{-1} \|g\| \cdot \|g^\perp\| = \frac{2}{\lambda} \|g\| \cdot \|g^\perp\|
\end{equation}


\noindent To express this in terms of the \( F^{-1/2} \)-norm, recall \textit{Lemma} \ref{lem:lemma2}:
\begin{align*} 
\|g\| &\leq \mu_{\max}^{1/2} \|F^{-1/2} g\|, &
\|g^\perp\| &\leq \mu_{\max}^{1/2} \|F^{-1/2} g^\perp\| \label{eq:mu-bound}
\end{align*}

\noindent Therefore:
\begin{equation}
|g^\top Q g^\perp| \leq \frac{2\mu_{\max}}{\lambda} \|F^{-1/2} g\| \cdot \|F^{-1/2} g^\perp\| \label{eq:cross-final}
\end{equation}

\vspace{1em}
\subsubsection{Bounding \( \beta^2 (g^\perp)^\top Q g^\perp  \)\\} 
\vspace{0.5em}
\noindent From Eq. \ref{eq:Q}:
\begin{equation}
    Q = (F + \lambda I)^{-1} F (F + \lambda I)^{-1}
\end{equation}

\noindent In the eigenbasis of \( F \) (symmetric and positive semi-definite), where \( F = \operatorname{diag}(\Lambda_1, \dots, \Lambda_n) \) and \(\Lambda_i\) is the eigenvalue, we can express \( Q \) as a diagonal matrix with entries:
\begin{equation}
    Q_i = \frac{\Lambda_i}{(\Lambda_i + \lambda)^2}
\end{equation}

\noindent Let \( g^\perp = \sum_i h_i u_i \), where \( \{u_i\} \) are the eigenvectors of \( F \), and \( h_i = u_i^\top g^\perp \). Then:
\begin{equation}
    (g^\perp)^\top Q g^\perp = \sum_i \frac{\Lambda_i}{(\Lambda_i + \lambda)^2} h_i^2
\end{equation}

\noindent This achieves its maximum at \( \Lambda_i = \lambda \), therefore:
\begin{equation}
    (g^\perp)^\top Q g^\perp  \leq \frac{1}{4\lambda} \sum_i h_i^2 = \frac{1}{4\lambda} \|g^\perp\|^2
\end{equation}

\noindent Recall \textit{Lemma} \ref{lem:lemma2}, we have:
\[
\|g^\perp\| \leq \mu_{\max}^{1/2} \|F^{-1/2} g^\perp\|
\]
Substitute into the bound:
\begin{equation}
    \beta^2 (g^\perp)^\top Q g^\perp \leq \frac{\beta^2}{4\lambda} \mu_{\max} \|F^{-1/2} g^\perp\|^2
\end{equation}

\noindent \textbf{Finally}, Eq. \ref{eq:KL-norm} becomes:

\begin{align}
\|F^{1/2} \Delta_{\text{FOP}}\|^2 
&\leq \eta^2 \Bigg[
\underbrace{g^\top Q g}_{\|F^{1/2} \Delta_{\text{KFAC}}\|^2} \nonumber \\
&+ 2\beta \frac{\mu_{\max}}{\lambda} \|F^{-1/2} g\| \cdot \|F^{-1/2} g_\perp\| \notag \nonumber \\
&\quad + \beta^2 \frac{\mu_{\max}}{4\lambda} \|F^{-1/2} g_\perp\|^2
\Bigg]
\end{align}

The KL-norm of an FOP update eventually decomposes into three terms: the K-FAC core term, which shrinks quadratically as \( \lambda^{-2} \), and two FOP-specific terms, a cross term and an orthogonal term, which shrink linearly as \( \lambda^{-1} \). Crucially, the FOP-specific terms are also scaled by \( \|F^{-1/2} g^\perp\| \), which is typically smaller than \( \|F^{-1/2} g\| \).

The analysis deliberately combines an exact expression for the core term with a worst-case bound for the orthogonal term to ensure safety even when the spectrum of \( g^\perp \) is unknown. In the large-damping training that K-FAC relies on for stability at big batch sizes, this bound guarantees that the FOP-specific terms remain at least one order of \( \lambda \) smaller than the core term.

This allows FOP to reduce \( \lambda \), preserving more curvature information while still respecting the KL trust-region. As a result, FOP achieves faster convergence in large-batch training without compromising stability.

\section{C. Experiments}
\subsection{C.1. Setup of CIFAR-10}
The training of ResNet-18~\cite{he2016deep} on the CIFAR-10~\cite{cifar} dataset serves as a fundamental experiment in the field of image classification. In this subsection, we compare the proposed FOP optimizer with several established baselines, including SGD with momentum (referred to as SGD~\cite{sgd}), AdamW~\cite{adamw}, and KFAC.
\noindent
We follow standard experimental settings and apply commonly used data augmentation techniques consisting of random cropping and horizontal flipping ~\cite{devries2017cutout}. All models are trained for 100 epochs.
For SGD, KFAC, and FOP, the initial learning rate $\alpha_0$ for batch size of $1024$ is selected via grid search over the set $\alpha \in \{10^{-2}, 5 \times 10^{-2}, 10^{-1}, 5 \times 10^{-1}, 1\}$. For AdamW, the grid is set to $\alpha \in \{10^{-4}, 5 \times 10^{-4}, 10^{-3}, 5 \times 10^{-3}, 10^{-2}\}$.
The update intervals for the curvature matrix and its inverse in both KFAC and FOP are synchronised, and we evaluate two base intervals: $\{10, 100, 200\}$. Each experiment is repeated with 5 different random seeds ($\{0, 1, 2,3,4\}$) to account for variability.
We use \texttt{ReduceLROnPlateau} as the learning rate scheduler, configured with a patience of 5 epochs, a decay factor of 0.1, a relative threshold of 0.0001, and \texttt{threshold\_mode} set to \texttt{rel}.

\subsection{C.2. Setup of ImageNet-100 experiment}

The implementation of T2T-ViT~\cite{Yuan_2021_ICCV} follows the original paper. We use the linear warmup strategy~\cite{goyal2017accurate} in the first 5 epochs for  AdamW, KFAC and FOP. The initial learning rate $\alpha_0$ is selected via grid search over $\alpha \in \{10^{-5}, 10^{-4}, \ldots, 10^{-1}\}$ for a batch size of $512$.
.The update intervals for the curvature matrix and inverse matrix correlating with KFAC and FOP are set to be $\{500, 1000\}$ for a batch size of 512. All models are trained for 100 epochs with three random seeding ($\{0, 1, 2\}$). AdamW, KFAC, and FOP use the cosine learning rate updating strategy and are set to be
\begin{align}
\label{eq:cosine}
\alpha_t = 0.001 
+ 0.5 &\times (\alpha_0 - 0.001) \nonumber \\
&\times \left(1 + \cos\left(\frac{2 \times 0.47 \times \pi \times t}{\text{max\_epoch}}\right)\right)
\end{align}


\subsection{C.3. Setup of ImageNet-1K experiment}


We implement \texttt{ResNet50} following the official PyTorch example~\cite{pytorch_examples_github}, based on the architecture proposed by He et al.~\cite{he2016deep}. For all optimizers (SGD, Adam, KFAC, and FOP), we apply a linear warm-up strategy~\cite{goyal2017accurate} during the first 5 epochs. For the KFAC and FOP optimizers, the update intervals for the curvature matrix and its inverse are set to $\{800, 1600, 2400\}$ when using a batch size of $1024$.
The total number of training epochs differs by optimizer: we train SGD and Adam for 80 epochs, while KFAC and FOP are trained for 50 epochs. SGD uses the cosine annealing learning rate schedule as described in Eq.~\ref{eq:cosine}. In contrast, KFAC and FOP adopt an exponential decay learning rate schedule given by:
\[
\alpha_t = \alpha_0 \times \left(1 - \frac{t}{\text{max\_epoch}}\right)^E
\]
where $t$ is the current epoch, and $E$ is the decay exponent, with $E \in \{4, 5, 6\}$. The initial learning rate $\alpha_0$ is selected via grid search over $\alpha \in \{10^{-5}, 10^{-4}, \ldots, 10^{-1}\}$ for a batch size of $1024$.

\begin{figure*}
  \centering
  \includegraphics[width=\linewidth]{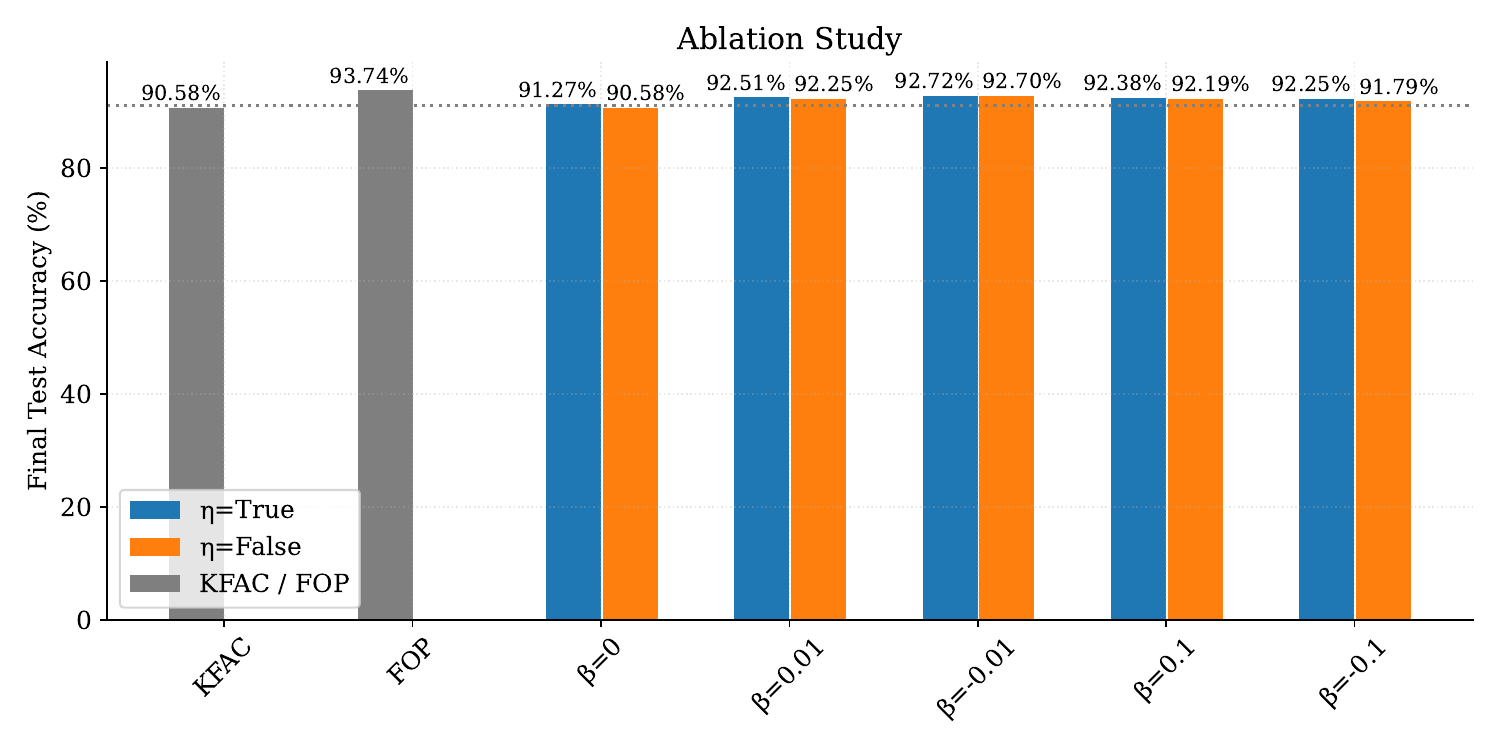}
  \caption{\textbf{Ablation of the scaling parameter $\eta$ on CIFAR-10.}
           Bars are grouped by optimiser KFAC (left cluster) and FOP (right cluster), and coloured by the value of the scaling term $\eta\in\{-0.1,-0.01,0,0.01,0.1\}$.
           Numbers above each bar give the Top 3 test accuracy averaged over three random seeds.}
  \label{fig:cifar10-eta-ablation}
\end{figure*}

\subsection{C.4. Additional experimental details and results}
\subsubsection{C.4.1 Damping ratio vs acc}

\begin{figure}[t]
  \centering
  \includegraphics[width=\linewidth]{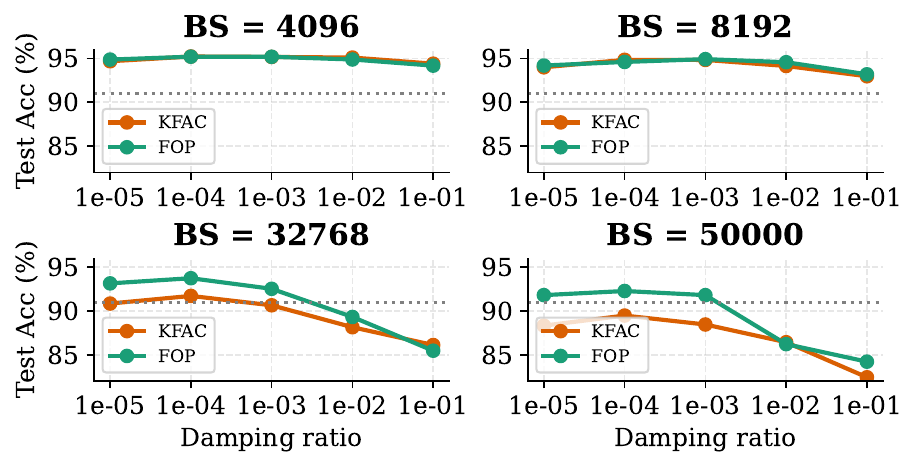}
  \caption{Best test accuracy vs.\ damping ratio for ResNet-18 on CIFAR-10, grouped by batch size. The dotted line represents the threshold of 91\%.}
  \label{fig:damping-cifar}
\end{figure}

\begin{figure}[t]
  \centering
  \includegraphics[width=\linewidth]{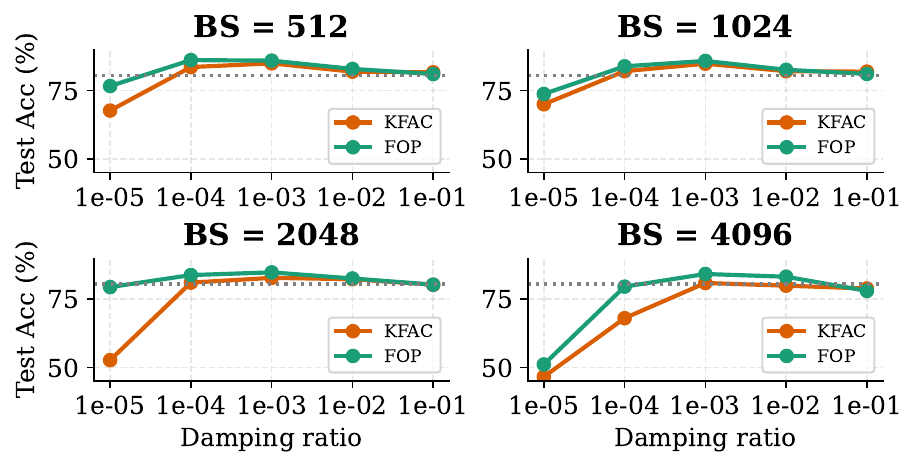}
  \caption{Best test accuracy vs.\ damping ratio for T2T-ViT on ImageNet-100, grouped by batch size. The dotted line represents the threshold of 80.9\%.}
  \label{fig:damping-vit}
\end{figure}

\begin{figure}[t]
  \centering
  \includegraphics[width=\linewidth]{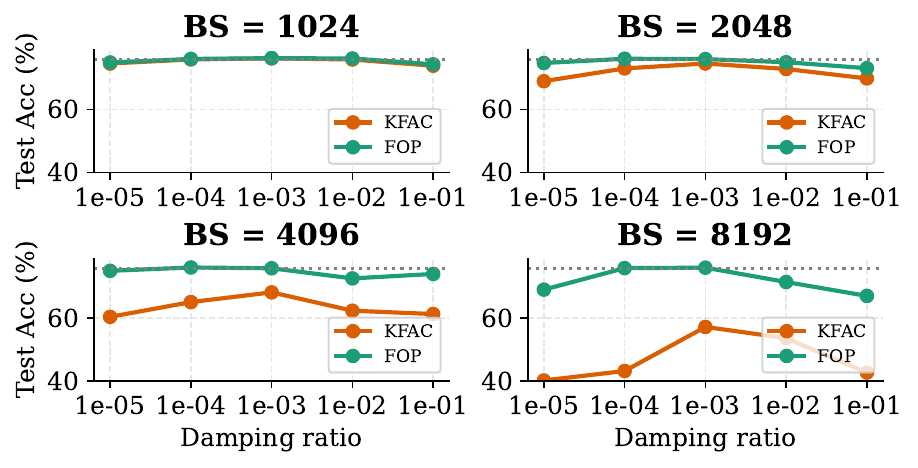}
  \caption{Best test accuracy vs.\ damping ratio for ResNet-50 on ImageNet-1K, grouped by batch size. The dotted line represents the threshold of 75.9\%.}
  \label{fig:damping-imagenet}
\end{figure}

Figure~\ref{fig:damping-cifar} presents results on CIFAR-10 with ResNet-18. Across all batch sizes ($2048$--$50000$), the proposed FOP optimiser consistently surpasses the 91\,\% test-accuracy threshold (dotted line) when the damping ratio $\lambda \in [10^{-5}, 10^{-3}]$. In contrast, KFAC fails to meet the cutoff whenever the damping is too aggressive ($\lambda \ge 10^{-2}$) and becomes unstable at $\lambda = 10^{-5}$. While both methods perform best around $\lambda = 10^{-3}$, FOP’s flatter curve indicates a much larger tolerance to hyper-parameter mis-specification.

\noindent
This pattern extends to transformer-based models as well. As shown in Figure~\ref{fig:damping-vit}, for T2T-ViT trained on ImageNet-100, FOP reliably exceeds the 80.9\,\% bar for all batch sizes and for all damping values except the extreme low end. In contrast, KFAC only reaches the target within the narrow range $\lambda \in [10^{-4}, 10^{-3}]$. Notably, FOP’s accuracy varies by less than one percentage point across the entire sweep from $\lambda = 10^{-5}$ to $10^{-1}$ at smaller batch sizes, highlighting its robustness even on architectures that include attention layers.
\noindent
A similar trend appears on the large-scale ImageNet-1K task with ResNet-50, as illustrated in Figure~\ref{fig:damping-imagenet}. Here, FOP again outperforms or matches KFAC across the board and remains above the 75.9\,\% accuracy threshold throughout the full sweep and across all batch sizes ($2048$--$8196$). KFAC’s performance, however, drops sharply for $\lambda \ge 10^{-2}$ and exhibits visible instability at the smallest damping, echoing the behaviour observed on CIFAR-10.
\noindent
Collectively, Figures~\ref{fig:damping-cifar}--\ref{fig:damping-imagenet} indicate that FOP provides a substantially broader and more stable operating range for the damping ratio compared to KFAC, across both convolutional and transformer-based architectures, and over a wide range of data scales. This increased robustness reduces the sensitivity to hyper-parameter selection and supports the viability of FOP as a reliable second-order optimisation method for large-batch training in diverse tasks.

\subsubsection{C.4.2 GPU memory }

Tables~\ref{tab:cifar-mem}--\ref{tab:imagenet1k-mem} report the peak GPU
memory usage for CIFAR-10/ResNet-18, ImageNet-100/T2T-ViT, and
ImageNet-1K/ResNet-50, respectively, along with the number of devices
used in each run.  
On CIFAR-10 (Table~\ref{tab:cifar-mem}), both SGD and AdamW exhibit the
lowest memory footprint, with usage scaling nearly linearly with batch
size.  KFAC incurs a moderate overhead, while FOP demands the highest
memory—up to 145\,GB for a batch of 16\,384 on a single GPU, as it needs to keep two gradients $g_1$ and $g_2$ under the same device. This vanishes when there are more than 1 device.  
For ImageNet-100 with T2T-ViT (Table~\ref{tab:vit-mem}), all three
methods operate near full device capacity (\(\approx\)150–170\,GB) 
at the smallest batch size, with FOP consistently exceeding the others
by 5–20\,GB.  
On ImageNet-1K with ResNet-50 (Table~\ref{tab:imagenet1k-mem}), memory
per GPU remains constant within each optimiser regardless of batch size,
since larger batches are distributed across more devices.  Here, SGD
requires \(\approx 88\)\,GB per GPU, KFAC \(\approx 98\)\,GB, and FOP
again matches KFAC closely except for the smallest batch where it peaks
at 153\,GB.
\vspace{-0.5em} 

\begin{table}[htbp]

\centering
\scriptsize
\renewcommand{\arraystretch}{0.5} 
\begin{tabular}{lrr}
\hline
\textbf{Optimizer} & \textbf{\# of GPU} & \textbf{Max Mem/GPU (GB)} \\
\hline
\multicolumn{3}{c}{\textbf{Batch Size: 2048}} \\
SGD   & 1 & 9.25 \\
AdamW & 1 & 9.30 \\
KFAC  & 1 & 14.51 \\
FOP   & 1 & 19.21 \\
\hline
\multicolumn{3}{c}{\textbf{Batch Size: 4096}} \\
SGD   & 1 & 18.40 \\
AdamW & 1 & 18.44 \\
KFAC  & 1 & 28.03 \\
FOP   & 1 & 37.20 \\
\hline
\multicolumn{3}{c}{\textbf{Batch Size: 8192}} \\
SGD   & 1 & 36.70 \\
AdamW & 1 & 36.74 \\
KFAC  & 1 & 55.08 \\
FOP   & 1 & 73.18 \\
\hline
\multicolumn{3}{c}{\textbf{Batch Size: 16384}} \\
SGD   & 1 & 73.29 \\
AdamW & 1 & 73.33 \\
KFAC  & 1 & 109.18 \\
FOP   & 1 & 145.16 \\
\hline
\multicolumn{3}{c}{\textbf{Batch Size: 32768}} \\
SGD   & 2 & 73.29 \\
AdamW & 2 & 73.38 \\
KFAC  & 2 & 108.31 \\
FOP   & 2 & 108.44 \\
\hline
\end{tabular}
\vspace{-0.5em}
\caption{Max GPU memory usage and number of GPUs for CIFAR-10 training with ResNet-18.\label{tab:cifar-mem}}
\end{table}

\begin{table}[H]
\centering
\scriptsize
\renewcommand{\arraystretch}{0.5}
\begin{tabular}{lrr}
\hline
\textbf{Optimizer} & \textbf{\# of GPU} & \textbf{Max Memory/GPU (GB)} \\
\hline
\multicolumn{3}{c}{\textbf{Batch Size: 512}} \\
AdamW & 1 & 145.41 \\
KFAC  & 1 & 153.28 \\
FOP   & 1 & 172.51 \\
\hline
\multicolumn{3}{c}{\textbf{Batch Size: 1024}} \\
AdamW & 1 & 151.10 \\
KFAC  & 1 & 158.85 \\
FOP   & 1 & 163.18 \\
\hline
\multicolumn{3}{c}{\textbf{Batch Size: 2048}} \\
AdamW & 1 & 148.54 \\
KFAC  & 1 & 156.86 \\
FOP   & 1 & 159.20 \\
\hline
\multicolumn{3}{c}{\textbf{Batch Size: 4096}} \\
AdamW & 1 & 142.95 \\
KFAC  & 1 & 151.20 \\
FOP   & 1 & 153.66 \\
\hline
\end{tabular}
\vspace{-0.5em}
\caption{Max GPU memory usage across optimizers and number of GPUs used per run for training ImageNet-100 with T2T-ViT.\label{tab:vit-mem}}
\end{table}

\vspace{-1em}

\begin{table}[H]
\centering
\scriptsize
\renewcommand{\arraystretch}{0.5} 
\begin{tabular}{lrr}
\hline
\textbf{Optimizer} & \textbf{\# of GPU} & \textbf{Max Memory/GPU (GB)} \\
\hline
\multicolumn{3}{c}{\textbf{Batch Size: 1024}} \\
SGD   & 1 & 87.89 \\
KFAC  & 1 & 110.31 \\
FOP   & 1 & 153.31 \\
\hline
\multicolumn{3}{c}{\textbf{Batch Size: 2048}} \\
SGD   & 2 & 87.99 \\
KFAC  & 2 & 98.04 \\
FOP   & 2 & 98.24 \\
\hline
\multicolumn{3}{c}{\textbf{Batch Size: 4096}} \\
SGD   & 4 & 87.99 \\
KFAC  & 4 & 98.04 \\
FOP   & 4 & 98.24 \\
\hline
\multicolumn{3}{c}{\textbf{Batch Size: 8192}} \\
SGD   & 8 & 87.99 \\
KFAC  & 8 & 97.89 \\
FOP   & 8 & 98.19 \\
\hline
\end{tabular}
\vspace{-0.5em}
\caption{Max GPU memory usage and number of GPUs used per run for training ImageNet-1K with ResNet-50.\label{tab:imagenet1k-mem}}
\end{table}

\subsection{C.4.3 Scaling efficiency}

\begin{figure}[h]
  \centering
  \includegraphics[width=\linewidth]{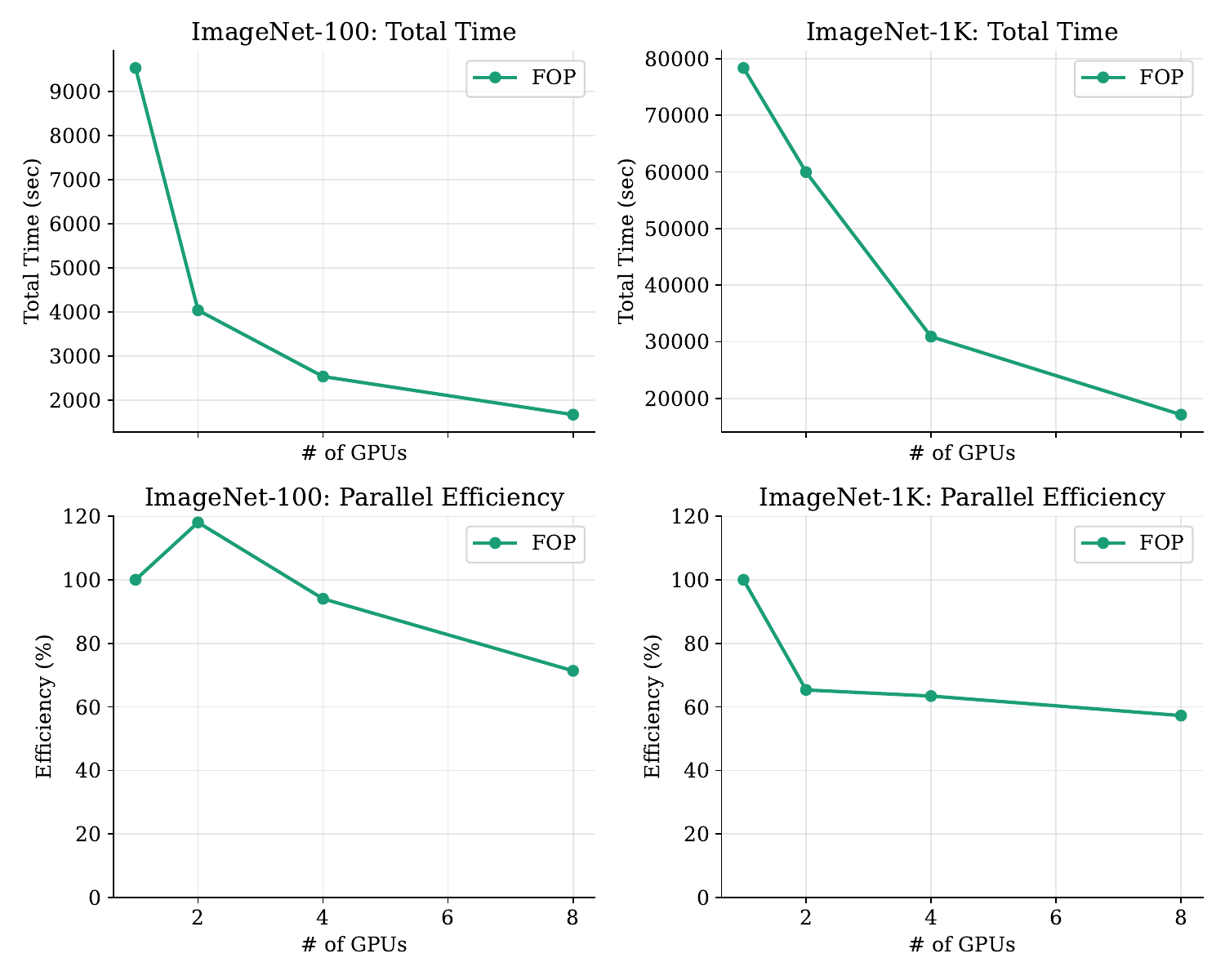}
  \caption{Strong-scaling results for FOP on ImageNet-1K.
           Top: wall-clock training time versus number of GPUs for
           ImageNet-100 (left) and ImageNet-1K (right).  
           Bottom: total GPU-time (\#GPUs × wall-clock) versus number of GPUs for the same two datasets.
           Dashed lines indicate ideal linear scaling.}
  \label{fig:imagenet-scaling}
\end{figure}

\section{D. Ablation Study of FOP}

\subsection{D.1. Effect of \(\beta\)}

An ablation study on CIFAR-10, shown in Figure~\ref{fig:cifar10-eta-ablation}, evaluates the effect of varying the momentum parameter $\beta$ in the FOP optimiser. Using ResNet-18 trained for 100 epochs (averaged over three seeds), the study compares different configurations of FOP against KFAC, which fixes $(\eta, \beta) = (1, 0)$.
FOP decouples the learning rate scale $\eta$ from the momentum factor $\beta$, allowing both to be fixed or adaptively updated. When $\eta$ is not adaptive (set to $1$), introducing $\beta$ consistently improves accuracy over KFAC. Further gains are observed when $\eta$ is adaptive, highlighting the benefit of dynamic learning rate scaling.
Overall, the results demonstrate that FOP’s flexibility in tuning or adapting $(\eta, \beta)$ leads to improved accuracy and robustness compared to fixed-setting baselines.

\section{E.  Comparison between Nvidia and AMD cards}

\subsection*{E.1. System Configuration}

\paragraph{AMD System (Training Runs)}
\begin{itemize}
  \item \textbf{Node Type:} Lenovo ThinkSystem SR685a V3
  \item \textbf{CPUs:} 2× AMD EPYC 9534 (64 cores each, 128 total cores, no SMT)
  \item \textbf{Max Frequency:} 3.72\,GHz, Frequency Boost enabled
  \item \textbf{Instruction Support:} AVX-512, AVX512-BF16, AVX512-VNNI, AVX512-VBMI
  \item \textbf{Cache:} L3 cache total 512\,MiB (16×32\,MiB)
  \item \textbf{GPUs:} 8× AMD MI300X
  \item \textbf{Software Stack:} PyTorch 2.6.0 + ROCm 6.2.4
  \item \textbf{ROCm Driver Version:} 6.2.1
\end{itemize}

\paragraph{NVIDIA System (Benchmarking)}
\begin{itemize}
  \item \textbf{Node Type:} Custom system
  \item \textbf{CPUs:} 2× Intel Xeon Platinum 8468 (48 cores each, 96 total cores)
  \item \textbf{Max Frequency:} 3.8\,GHz
  \item \textbf{Instruction Support:} AVX-512, AVX-VNNI, AMX (Tile, INT8, BF16)
  \item \textbf{GPUs:} 8× NVIDIA H100 (80\,GB HBM3)
  \item \textbf{Software Stack:} PyTorch 2.8.0.dev20250413+cu126, CUDA 12.9
  \item \textbf{Driver Version:} 575.57.08
\end{itemize}

\begin{figure*}
  \centering
  \includegraphics[width=\textwidth]{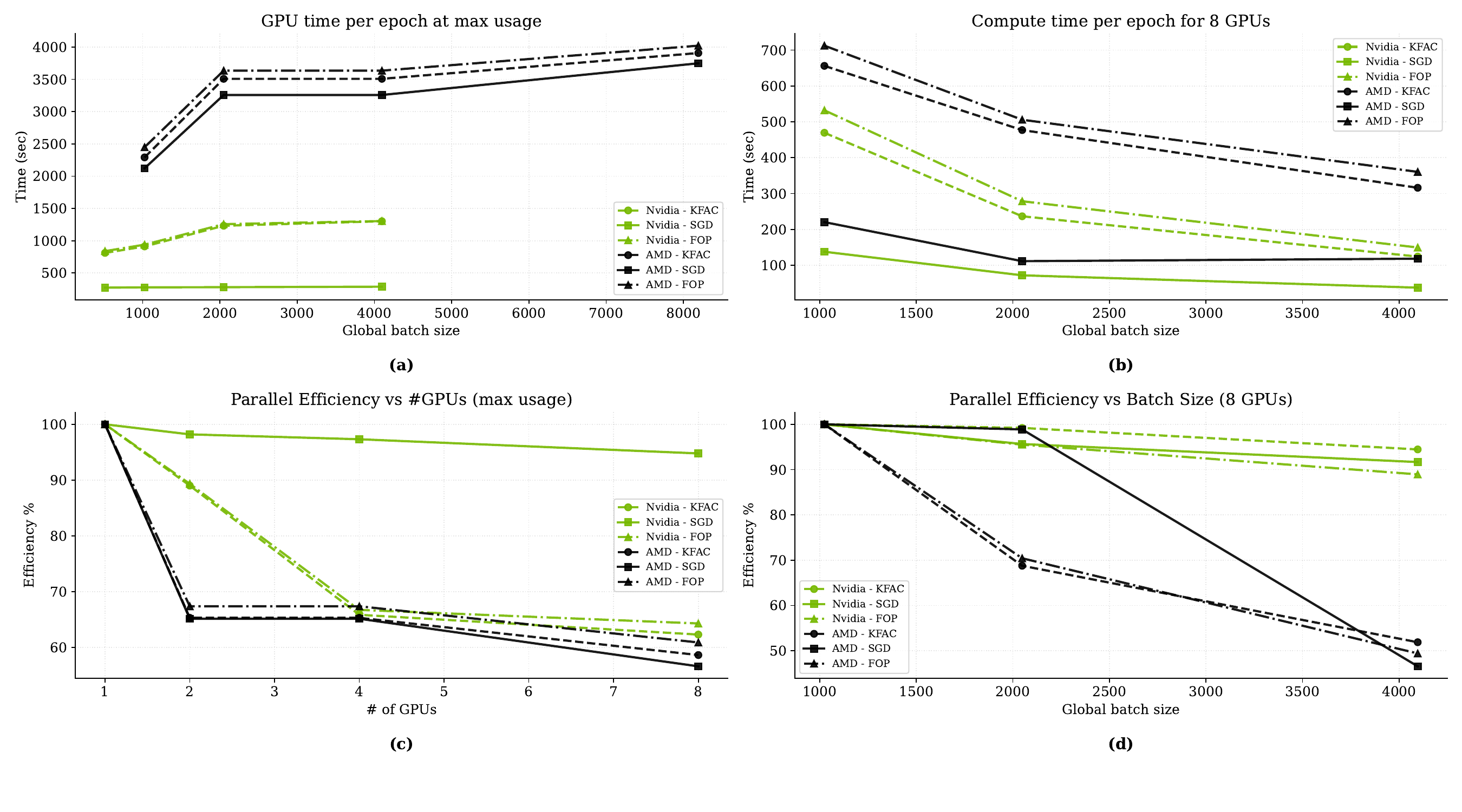}
  \caption{
    Training time and parallel efficiency comparisons across different methods (SGD, KFAC, FOP) and hardware (NVIDIA H100 and AMD MI300X). 
    (a) Total GPU time per epoch scaled by the number of GPUs used, representing overall computational effort. 
    (b) Raw wall-clock time per epoch using 8 GPUs. 
    (c) \textbf{Parallel efficiency vs number of GPUs}, calculated as \( \text{Efficiency} = \frac{T_1}{n T_n} \), where \( T_1 \) is the time with the smallest GPU count and \( T_n \) the time using \( n \) GPUs. 
    (d) \textbf{Efficiency vs global batch size on 8 GPUs}, defined as \( \text{Efficiency} = \frac{T_{b_0} \cdot b_0}{T_b \cdot b} \), measuring how well increased batch sizes are utilized on fixed hardware.
  }
  \label{fig:training-efficiency}
\end{figure*}







\subsection*{E.2. Benchmarking Results on ImageNet-1k with ResNet-50}

To evaluate both the scalability and efficiency of different optimization methods across hardware platforms, we present four subplots in Figure~\ref{fig:training-efficiency}. 
Subplot (a) shows the \emph{total GPU time per epoch}, which accounts for both the per-step time and the number of GPUs used, reflecting the total computational cost. 
Subplot (b) illustrates the \emph{raw wall-clock time per epoch} when training with 8 GPUs.

In subplot (c), we measure the \textbf{strong scaling parallel efficiency}, defined as
\[
\text{Efficiency}_{\#\text{GPUs}} = \frac{T_1}{n \cdot T_n} \times 100\%,
\]
where \( T_1 \) is the training time using the smallest GPU count (typically one GPU), \( T_n \) is the time when using \( n \) GPUs, and \( n \) is the number of GPUs. This metric indicates how effectively the training process scales with added hardware resources.

Subplot (d) evaluates the \textbf{efficiency with respect to batch size at fixed GPU count} (8 GPUs), defined as
\[
\text{Efficiency}_{\text{batch}} = \frac{T_{b_0} \cdot b_0}{T_b \cdot b} \times 100\%,
\]
where \( T_{b_0} \) is the epoch time at a reference batch size \( b_0 \), and \( T_b \) is the epoch time for a new batch size \( b \). This quantifies how well larger batch sizes maintain per-sample throughput on the same hardware.

Together, these plots provide insights into both inter-GPU scaling behavior and intra-GPU efficiency across diverse training methods and compute platforms.

\end{document}